\documentclass[10pt,twocolumn,letterpaper]{article}

\usepackage{cvpr}              

\usepackage[dvipsnames]{xcolor}

\usepackage[accsupp]{axessibility}
\definecolor{cvprblue}{rgb}{0.21,0.49,0.74}
\usepackage[pagebackref,breaklinks,colorlinks,citecolor=cvprblue]{hyperref}
\usepackage{cuted}
\usepackage{capt-of}
\usepackage{bbm}
\usepackage{caption}
\usepackage{subcaption}
\usepackage{makecell}
\usepackage{float}
\usepackage{amsmath}
\usepackage{amssymb}
\usepackage[notransparent]{svg}

\newcommand{\authorsep}{\hspace{8pt}}
\newcommand{\affiliationsep}{\hspace{8pt}}
\newcommand{\EqCont}{${}^{*}$}
\newcommand{\AffI}{${}^1$}

\newcommand{\AffIandII}{${}^{1,2}$}
\newcommand{\Aff}{${}^{1,3}$}

\def\paperID{11646} 
\def\confName{CVPR}
\def\confYear{2024}

\title{Zero-Painter: Training-Free Layout Control for Text-to-Image Synthesis}

\author{Marianna Ohanyan\AffI\EqCont \authorsep
Hayk Manukyan\AffI\EqCont \authorsep
Zhangyang Wang\AffIandII \\
Shant Navasardyan\AffI \authorsep
Humphrey Shi\Aff \\
\small${}^1$Picsart AI Research (PAIR) \affiliationsep
\small${}^2$UT Austin \affiliationsep
\small${}^3$Georgia Tech
\\{\small \textbf{\url{https://github.com/Picsart-AI-Research/Zero-Painter}}}
}

\begin{document}

\maketitle
\begin{strip}
    \centering
    \includegraphics[width=\textwidth]{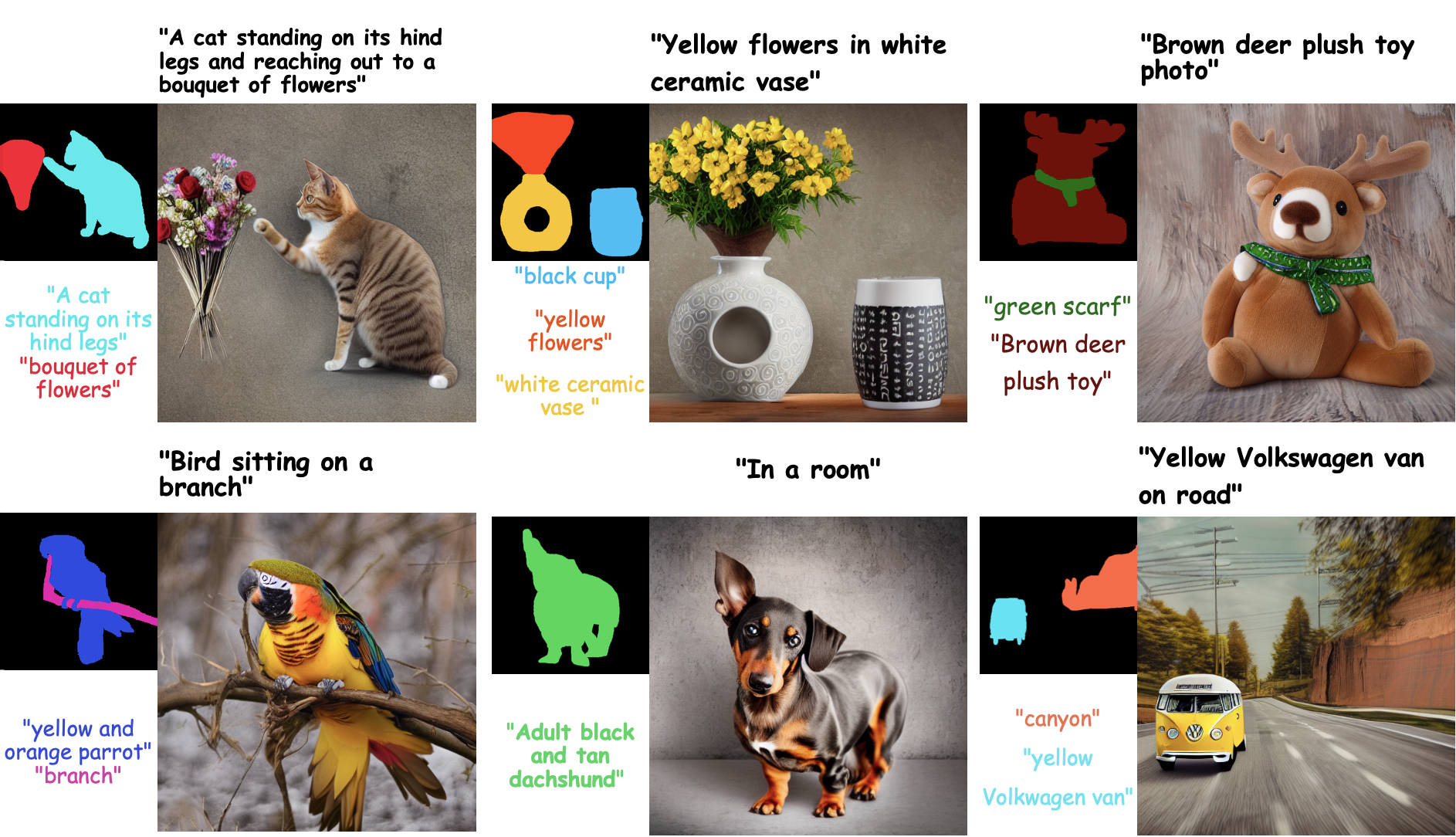} 
    \captionof{figure}{Embark on a visual journey with Zero-Painter: a novel training-free framework for layout-conditional text-to-image generation. This new pipeline brings images to life using object masks and individual descriptions, seamlessly fused with a powerful global text prompt.}
    \label{fig:teaser_img}
\end{strip}
\def\thefootnote{*}\footnotetext{Equal contribution.}\def\thefootnote{\arabic{footnote}}

\begin{abstract}

We present Zero-Painter, a novel training-free framework for layout-conditional text-to-image synthesis that facilitates the creation of detailed and controlled imagery from textual prompts. Our method utilizes object masks and individual descriptions, coupled with a global text prompt, to generate images with high fidelity. Zero-Painter employs a two-stage process involving our novel Prompt-Adjusted Cross-Attention (PACA) and Region-Grouped Cross-Attention (ReGCA) blocks, ensuring precise alignment of generated objects with textual prompts and mask shapes. Our extensive experiments demonstrate that Zero-Painter surpasses current state-of-the-art methods in preserving textual details and adhering to mask shapes.

\end{abstract}
\section{Introduction}
\label{sec:intro}

Recent innovations in generative AI have revolutionized the creative landscape, allowing the generation of strikingly realistic images \cite{Dalle2,stable_diff,imagen_paper} or videos \cite{videoworldsimulators2024, Blattmann_Dockhorn,text2video-zero,StreamingT2V} from text.
However, crafting detailed prompts to guide every aspect of an image can be cumbersome and time-intensive. 
Furthermore, traditional text-to-image models often falter when faced with intricate prompts that 
describe multiple objects and their respective attributes. 
To address these challenges, layout-conditional text-to-image models have been developed 
that leverage additional inputs such as segmentation masks \cite{eDiff,gligen,multidiffusion,SpaText} 
or bounding boxes \cite{gligen,no_token_left_behind,Yang2022} together with text. This approach facilitates the creation of images with precise attributes, 
granting artists and designers granular control over the visual components.

Early iterations of layout-conditional text-to-image methods \cite{gan_panoptic,make_a_scene}, 
employing GANs \cite{goodfellow2014GAN} and diffusion models \cite{DDPM_paper,DDIM_paper}, 
achieved remarkable results using a closed vocabulary. 
However, their reliance on fixed class labels restricted their ability to prompt 
free-form attributes for the object in the layout. 
The introduction of GLIGEN \cite{gligen} marked a significant advancement 
with its open-vocabulary and a multitude of new control mechanisms,
including bounding box and text pairs, keypoints, edges, depth, and class-based segmentation maps. 
Later eDiff-I \cite{eDiff} introduced the Paint-With-Words approach, 
allowing open-vocabulary prompting with free-form mask control.
Subsequently, MultiDiffusion  \cite{multidiffusion} was introduced, 
capable of processing local prompts independently from the global prompt, adding flexibility.
While these methods can generate visually convincing and prompt-aligned results, they are not always capable of keeping the shapes of the objects inside the mask.
Despite this innovation, the absence of explicit mask conditioning often resulted in discrepancies between the shapes of generated objects and the provided masks.

To overcome these challenges, we present Zero-Painter, 
an innovative training-free method for layout-conditional text-to-image synthesis. 
It generates images from object masks and individual descriptions, 
alongside a global text prompt, as showcased in Fig. \ref{fig:teaser_img}. 
Our process is bifurcated into two stages: initially, we generate individual objects, 
each endowed with unique attributes, using our Prompt-Adjusted Cross-Attention (PACA) module. 
These objects are then seamlessly integrated into a single scene through our Region-Grouped Cross-Attention (ReGCA) block. 
This ensures the generated objects not only align with the prompts but also conform to the shapes of the provided masks. 
Through rigorous testing, we have found that Zero-Painter surpasses state-of-the-art methods, 
particularly in maintaining the textual integrity of individual objects
and adhering to the shapes of the given masks.

In summary, our contributions are as follows:
\begin{itemize}
    \item We introduce Zero-Painter, a novel training-free framework for layout-conditional text-to-image synthesis, enabling the generation of objects with specified shapes and distinct attributes.
    \item We unveil the Prompt-Adjusted Cross-Attention (PACA) and Region-Grouped Cross-Attention (ReGCA) blocks, which significantly improve the shape fidelity and characteristic preservation of the generated objects.
    \item Comprehensive experiments validate Zero-Painter's superiority over existing state-of-the-art methods, as evidenced through both quantitative and qualitative comparisons.
\end{itemize}
\section{Related Work}

\subsection{Text-to-Image Generation}
Recently, significant advancements have occurred in the field of text-to-image generation.
Approaches based on Generative Adversarial Networks (GANs)  \cite{goodfellow2020generative,zhang2017stackgan,xu2018attngan,qiao2019mirrorgan} have yielded promising results in constrained domains. 
With the rise of transformer  \cite{vaswani2017attention} models, 
zero-shot open-domain models were introduced. 
Notably, both Dall-E  \cite{Dalle_paper} and VQ-GAN  \cite{VQ_GAN_paper} propose a two-stage approach. 
Initially, they employ a discrete Variational Autoencoder (VAE)  \cite{kingma2013auto,razavi2019generating} to discover a comprehensive semantic space. 

Later, Parti  \cite{parti_paper} illustrates the practicality of expanding autoregressive models in terms of scalability.

With the introduction of Diffusion-based models  \cite{stable_diff}, the quality of text-to-image generation has significantly improved. 
DALL-E 2  \cite{Dalle2_paper} utilizes CLIP  \cite{CLIP_paper}  for the text-to-image mapping process through diffusion mechanisms and trains a CLIP decoder. 
Furthermore, Imagen  \cite{imagen_paper} leverages large pre-trained language models like T5 on textual data  \cite{raffel2019exploring}, achieving superior alignment between images and text, as well as enhanced sample fidelity.
Lastly, eDiff-I \cite{eDiff} employs an expert-based approach, with different expert models handling generation at various timestep ranges.

\subsection{Layout-to-Image Generation}
Past research focused on image generation from structured layouts with fixed classes for content control  \cite{spade,pix2pix2017,object_centric,he2021context,ma2020attributeguided,goel2023pair,xu2023prompt}. 
CLIP  \cite{clip} marked a paradigm shift by introducing zero-shot learning and enabling a transition from fixed to free-form text control. 
Recent advancements, including No-token-left-behind \cite{no_token_left_behind} and Gligen \cite{gligen}, 
proposed methods incorporating free-form text and bounding boxes,
while Make-A-Scene \cite{make_a_scene} presented an innovative approach with a fixed set of labels but free-form mask-based control. 
Later approaches like Spa-text \cite{SpaText}, eDiff-I \cite{eDiff}, and MultiDiffusion \cite{multidiffusion} combine elements of free-form text and masks to broaden the scope of image generation.

\subsection{Text-Guided Image Inpainting} 

The problem of image inpainting is known in the community and has been tackled in numerous works: 
 \cite{goodfellow2020generative,gated_conv,oninon_conv,hifill,rfr,co-mod-gan,LaMa,sh-gan,cm-gan,partial_conv,gated_conv,oninon_conv,manukyan2023hd}.
With the recent rise of text-guided generative models, the text-guided version of image-inpainting has become relevant.
Research in this field is progressing rapidly, and works like SmartBrush \cite{xie2022smartbrush}, Imagen Editor \cite{wang2022imagen}, and Uni-paint \cite{unipaint} 
present significant advancements.
Most notably, Stable Inpainting  \cite{stable_diff} is a modification of the Stable Diffusion model, that is fine-tuned for text-guided inpainting.
The base model is enhanced by concatenating the input image and inpainting mask as additional conditioning to the UNet model's latent input.
The weights of the additional channels have been initialized with zeros, and fine tuned on the LAION  \cite{LAION} dataset using randomly generated inpainting masks.

\section{Method}

In this section we introduce Zero-Painter, a training-free layout-conditional text-to-image generation framework. 
First, we provide a brief background on diffusion models, focusing on Stable Diffusion (SD) \cite{stable_diff}.
Then, we present an overview of Zero-Painter, our two-stage pipeline that encompasses Single Object Generation (SOG) and Comprehensive Composition (CC). 
We delve into each stage, highlighting the design of Prompt-Adjusted Cross-Attention (PACA) and Region-Grouped Cross-Attention (ReGCA) layers for improved shape alignment and characteristic preservation.


\subsection{Stable Diffusion}
\label{sec:method:stable_diffusion}

Stable Diffusion \cite{stable_diff} is an LDM that works in the latent space of VQ-VAE \cite{VQ_VAE_paper} (or VQ-GAN \cite{VQ_GAN_paper} for the original LDM).
During the diffusion process, Gaussian noise is iteratively added to the input latent tensor $x_0\in\mathbb{R}^{h\times w\times c}$,
such that the conditional distribution $q(x_t|x_{t-1})$ is:
\begin{equation}
    q(x_t|x_{t-1}) = \mathcal{N}(x_t; \sqrt{1-\beta_t}x_{t-1}, \beta_t I), \;
    t = 1,..,T
\end{equation}
where $\{\beta_t\}_{t=1}^{T}$ are hyperparameters, and $T$ is large enough that $x_T$ becomes very close to $\mathcal{N}(0,I)$.
By unraveling the process and denoting $\alpha_t = \prod_{i=1}^{t}(1-\beta_i)$ one can get a direct formula for $x_t$:
\begin{equation}\label{eq:ddpm_forward}
    x_{t} = \sqrt{\alpha_t} x_0 + \sqrt{1-\alpha_t} \epsilon , \; \epsilon \sim N(0,1).
\end{equation}
The objective of SD is to learn a backward process
\begin{equation}
    p_\theta(x_{t-1}|x_t) = \mathcal{N}(x_{t-1};\mu_\theta(x_t,t),\Sigma_\theta(x_t,t)),
\end{equation}
where $\mu_\theta(x_t,t), \Sigma_\theta(x_t,t)$ are parametric learnable functions (for simplicity usually $\Sigma_\theta(x_t,t) = \mbox{diag}(\sigma^2)$ for a hyperparameter $\sigma$).
Later, sampling $x_T \sim \mathcal{N}(0,I)$ and performing the backwards process for $t=T,\ldots,1$ allows the generation of valid images,
where final latent $x_0$ has to be decoded using the decoder $\mathcal{D}(x_0)$ of the VAE.

An alternative deterministic sampling approach was proposed in \cite{DDIM_paper} called DDIM:
\begin{equation}\label{eq:method:ddim1}
    \begin{split}
        x_{t-1} = \sqrt{\alpha_{t-1}}\left(\frac{x_t - \sqrt{1-\alpha_t}\epsilon^t_{\theta}(x_t)}{\sqrt{\alpha_t}}\right) + \\
        \sqrt{1-\alpha_{t-1}}\epsilon^t_{\theta}(x_t), \quad t=T,\ldots,1, 
    \end{split}
\end{equation}
where
\begin{equation}
    \epsilon^t_{\theta}(x_t) = \frac{\sqrt{1-\alpha_t}}{\beta_t}x_t + 
    \frac{(1-\beta_t)(1-\alpha_t)}{\beta_t}\mu_\theta(x_t,t).
\end{equation}

For Stable Diffusion the function $\epsilon^t_{\theta}(x_t,\tau)$ is directly predicted using a UNet.
For text-to-image generation the UNet is altered by adding Cross-Attention layers, 
and an additional input textual prompt $\tau$ is provided:
\begin{equation}
    \begin{split}
        x_{t-1} = \sqrt{\alpha_{t-1}}\left(\frac{x_t - \sqrt{1-\alpha_t}\epsilon^t_{\theta}(x_t,\tau)}{\sqrt{\alpha_t}}\right) + \\
        \sqrt{1-\alpha_{t-1}}\epsilon^t_{\theta}(x_t,\tau), \quad t=T,\ldots,1.
    \end{split}
\end{equation}

\subsection{Zero-Painter}\label{sec:method:overview}

\begin{figure*}
    \centering
    \includegraphics[width=\textwidth]{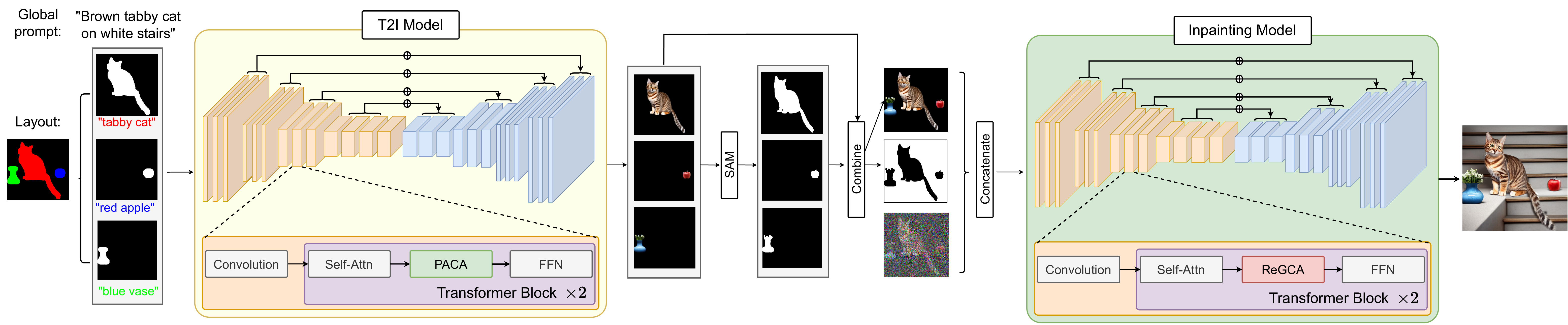} 
    \caption{Optimization-Free Two-Stage Pipeline for Zero-Shot Image Composition: (a) In the first stage, we focus on single object generation, leveraging the innovative Prompt-Adjusted Cross-Attention (PACA) layer. (b) Moving to the comprehensive composition stage, we introduce the Region-Grouped Cross-Attention (ReGCA) block, facilitating seamless and dynamic composition of generated objects.}
    \label{fig:arch-overview}
\end{figure*}

The problem of layout-conditional text-to-image generation can be formulated as follows:
Given a layout as a set of binary masks $M_i\in\{0,1\}^{H\times W}, \; i=1,\ldots,n$, indicating the shapes and positions of individual objects in a desired image, 
with corresponding textual prompts $\tau_i$  describing each object separately, 
as well as a global prompt $\tau_\text{global}$ describing the image in its entirety,
the goal is to generate an output image $I\in\mathbb{R}^{H\times W}$ matching $\tau_\text{global}$ while containing the objects following the shapes and positions of the layout $\{M_i\}_{i=1}^n$, and the prompts $\{\tau_i\}_{i=1}^n$.

To this end, Zero-Painter introduces an optimization-free two-stage pipeline, 
enabling the independent generation of objects followed by their seamless composition into a single image. 
This two-stage approach gives an advantage of utilizing the whole capacity of a diffusion model on single object generation resulting in better shape and characteristics alignment than current one-stage methods (see Fig. \ref{fig:qualitative-comparizon}).

During the first stage, we leverage PACA layer to individually generate images 
$I_i, \; i=1,\ldots, n$, each containing a single object on a flat background that
follows the shape/position of the binary mask $M_i$ and matches the description $\tau_i$ (see \cref{fig:arch-overview}).
The use of flat backgrounds aids in the easy identification and segmentation of the generated objects, especially if they differ slightly from the original mask.

The second stage takes the generated images $I_1, \ldots, I_n$, 
and combines them according to the global prompt $\tau_{global}$ and the individual mask-prompt pairs $(M_i, \tau_i)$.
To coherently combine the generated objects, we first separate them from their backgrounds 
by utilizing the Segment Anything Model (SAM) \cite{sam} and obtaining new object masks $M'_i\in\{0,1\}^{H\times W}$.
Note that using the initial mask $M_i$ for separating the foreground object in $I_i$ may cause undesirable object cuts
if the generated object shape has a slight mismatch with the given layout mask $M_i$ (see Fig. \ref{fig:sog-example}).
Then we leverage Stable Inpainting \cite{stable_diff} modified with our Region-Grouped Cross-Attention (ReGCA) module 
to seamlessly combine the generated objects and fill the background region indicated by the mask $M'_{bg} = (1-M'_1)\odot\ldots\odot(1-M'_n)$. 
The inpainting is done with the textual guidance of $\tau_{global}$.

\subsection{Single Object Generation (SOG)} \label{sec:method:sog}

The goal of the SOG stage is to produce an image from a binary mask $M_i$ and textual description $\tau_i$, 
ensuring the generated object's shape matches $M_i$ and its description matches $\tau_i$.
The rest of the image is filled with a flat background for easier separation of the generated objects later.

We use a pre-trained Stable Diffusion (SD) \cite{stable_diff} model to generate an image with a specified pair ($M_i$,$\tau_i$).
\cite{eDiff} shows that the high timesteps in the diffusion process are mostly responsible for creating the object silhouette,
while later steps create details and refine the object.
Since in our case the target shape is known and described by $M_i$, 
we start the diffusion backward process from an intermediate timestep $T' < T$,
and provide the shape information through a starting latent $x_{T'}$ obtained using the mask $M_i$.
To construct $x_{T'}$ we consider two factors: 
$(i)$ the background of the final image should be of a flat color; 
$(ii)$ the foreground object should be constrained to $M_i$.
We first obtain the latent code of the flat background corresponding to timestep $T'$
by applying noise on a latent encoding of a constant black image:

\begin{equation}\label{eq:method:flat_latent}
    x^\text{flat}_{T'} = \sqrt{\alpha_{T'}} \mathcal{E}(I^\text{flat}) + \sqrt{1-\alpha_{T'}}\epsilon,
\end{equation}
where $I^\text{flat}$ is the constant black image, $\mathcal{E}()$ is the VAE encoder
and $\epsilon \sim \mathcal{N}(\mathbf{0}, \mathbf{1})$ is sampled from the standard gaussian distribution.

We initialize the foreground region $M_i$ of $x_{T'}$ from a sampled  Gaussian noise $\epsilon \sim N(\textbf{0},\textbf{1})$.
While this deviates from the value expected according to \cref{eq:ddpm_forward}, 
we find that providing any specific value for $x_0$ may encourage the model to generate an image similar to $x_0$,
introducing unnecessary characteristics.
Furthermore, for sufficiently high values of $T'$ the noise to signal ratio is so high, 
that $\epsilon$ becomes a good enough approximation, and the deviation remains within the tolerance of the model. 
To further mitigate potential quality loss, 
we add a refinement sub-stage to object composition stage of Zero-Painter, 
aiming to correct any remaining errors in the generated image.

In summary, the starting latent $x_{T'}$ for the single object generation stage is defined as 
\begin{equation}\label{eq:method:starting_latent}
    x_{T'} = (1 - M_i) \odot x^\text{flat}_{T'} + M_i \odot \epsilon,
\end{equation}
where $M_i$ is the mask, and $\epsilon \sim \mathcal{N}(\textbf{0},\textbf{1})$ is randomly sampled from the standard gaussian distribution.

After we get the initial latent $x_{T'}$ we apply the DDIM backward process for $t=T',\ldots,1$ by leveraging SD, 
enhanced with our Prompt-Aware Cross-Attention (PACA) layers designed for the individual object shape alignment with $M_i$ and coherence with the prompt $\tau_i$:
\begin{equation}
\begin{split}
    x^\text{pred}_0(t) = \frac{x_t - \sqrt{1-\alpha_t}\epsilon^t_{\theta}(x_t)}{\sqrt{\alpha_t}}, \\
    x_{t-1} = \sqrt{\alpha_{t-1}}x^\text{pred}_0(t) + \sqrt{1-\alpha_{t-1}}\epsilon^t_{\theta}(x_t),
\end{split}
\end{equation}
where $\epsilon^t_{\theta}()$ is SD augmented with PACA layers.

To further ensure that the object does not extend outside of the masked area,
we blend the noised latent of the flat background $x^\text{flat}_t$ and the predicted $x_t$ at every timestep  $t=T', \ldots, 0$:
\begin{equation}
    x_t = M_i \odot x_t   + (1 - M_i) \odot x^\text{flat}_t,
\end{equation}

\subsubsection{Prompt-Aware Cross-Attention (PACA)}
\label{sec:method:sog-cross-att}

\begin{figure}
    \centering
    \includegraphics[width=0.3\linewidth]{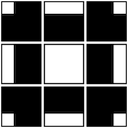}
    \includegraphics[width=0.3\linewidth]{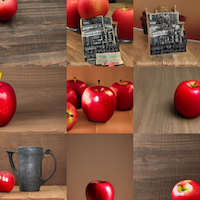}
    \caption{Effect of the SOT token. The similarity with the SOT token has been increased during text-to-image generation (at every step) in the non-white areas of the masks (left side). Prompt: "photo of a red apple, centered".}
    \label{fig:sim-apples}
\end{figure}

\begin{figure}
    \centering
    \begin{subfigure}[c]{0.25\linewidth}
        \includegraphics[width=\linewidth]{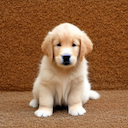}\subcaption{Output}
    \end{subfigure}
    \begin{subfigure}[c]{0.25\linewidth}
        \includegraphics[width=\linewidth]{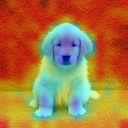}\subcaption{SOT Token}
    \end{subfigure}
    \begin{subfigure}[c]{0.25\linewidth}
        \includegraphics[width=\linewidth]{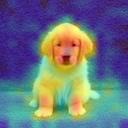}\subcaption{Other Tokens}
    \end{subfigure}
    
    \caption{Similarity of the SOT token vs all other tokens combined.}
    \label{fig:sim-puppy}
\end{figure}

\begin{figure}
    \centering
    \includegraphics[width=0.7 \linewidth]{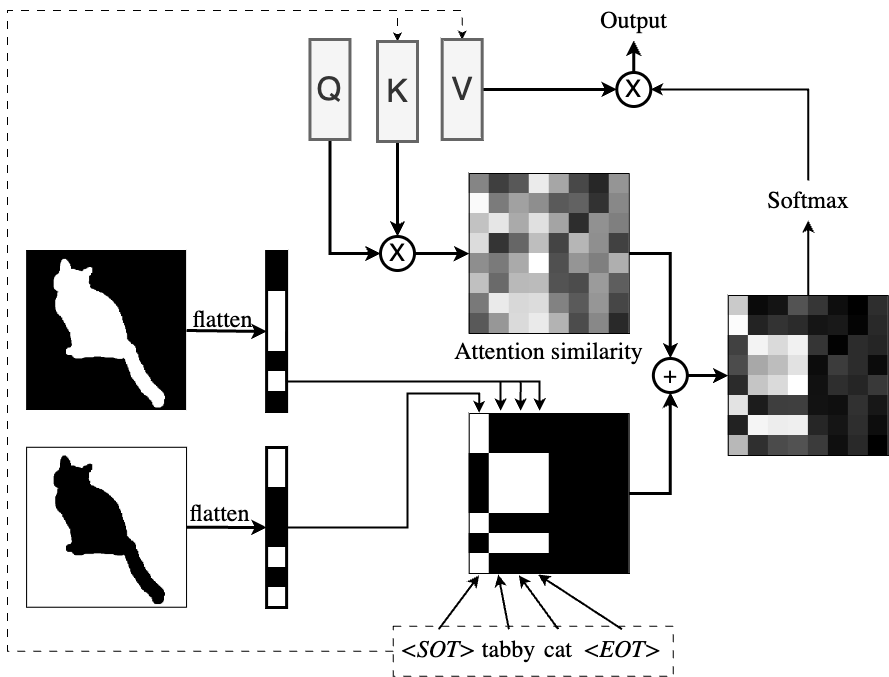} 
    \caption{Overview of Prompt-Aware Cross-Attention(PACA) during the Invdividual Object Generation stage.}
    \label{fig:sog}
\end{figure}

The PACA layer (Fig. \ref{fig:sog}) plays a crucial role in the SOG process.
It ensures accurate object generation based on textual descriptions and masks, while preventing generation outside the masked area.
For our PACA layer we employ a mechanism similar to eDiff-I \cite{eDiff},
i.e. by modifying the cross-attention similarity matrix $S$.
To encourage the generation of the object inside the masked area, 
we increase similarity values of queries $q_j, j \in M_i$ corresponding to the masked area $M_i$,
with the keys of all prompt tokens excluding the SOT.
We exclude the SOT since we noticed that during vanilla text-to-image generation
increasing the similarity of a pixel with the SOT token results in the output pixel becoming a generic background pixel (see \cref{fig:sim-apples,fig:sim-puppy}).
For the same reason, we increase the similarity values of non-masked pixel $q_j, j \notin M_i$ with the SOT token.
Therefore, the similarity matrix $S$ of selected cross attention layers is modified as follows:

\begin{equation}
    S_j' = 
    \begin{cases} 
        S_j + w_t \sum_{k = 1}^{N}{\mathbbm{1}_k}  & \text{if } j \in M_i \\
        S_j + w_t \mathbbm{1}_0 & \text{otherwise}
    \end{cases}
\end{equation}

where $S_j$ is the column of the similarity matrix corresponding to pixel with index $j$,  $\mathbbm{1}_k = [0 ... 1 ... 0]$  is an indicator vector and $N$ is the index of the EOT token, accordingly, index 0 is the SOT token.
Inspired by \cite{eDiff} we choose $w_t = w' \log(1 + \sigma_t)\max{(QK^T)}$, where $\sigma_t$ is the noise-to-signal ratio and $w'$ is a hyperparameter.

\subsection{Comprehensive Composition (CC)} \label{sec:method:composition}

\begin{figure}
    \centering
    \includegraphics[width=0.22\linewidth]{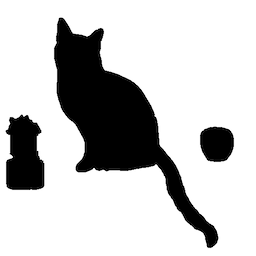} 
    \includegraphics[width=0.22\linewidth]{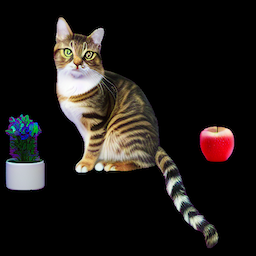} 
    \includegraphics[width=0.22\linewidth]{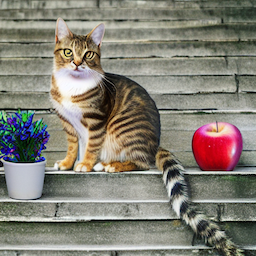} 
    \includegraphics[width=0.22\linewidth]{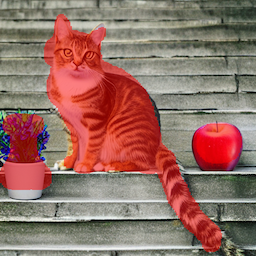} 
    \caption{Left to right: the cobined mask used for inpainting, the combined input image, the resulting comprehensive composition and an overlay with original input layout. }
    \label{fig:composition-example}
\end{figure}
The CC phase aims to combine all previously generated objects into a single image
that fits the description of the global prompt $\tau_\text{global}$. 
In \ref{sec:method:obj-seg-sam} we describe how to perform object segmentation for each individual object. 
In \ref{sec:method:Inpaint} - the process of inpainting the background region and 
in \ref{sec:method:ReGCA} - details concerning Region-Grouped Cross-Attention (ReGCA).
\subsubsection{Object Segmentation}
\label{sec:method:obj-seg-sam}
\begin{figure}
    \centering
    \includegraphics[width=0.22\linewidth]{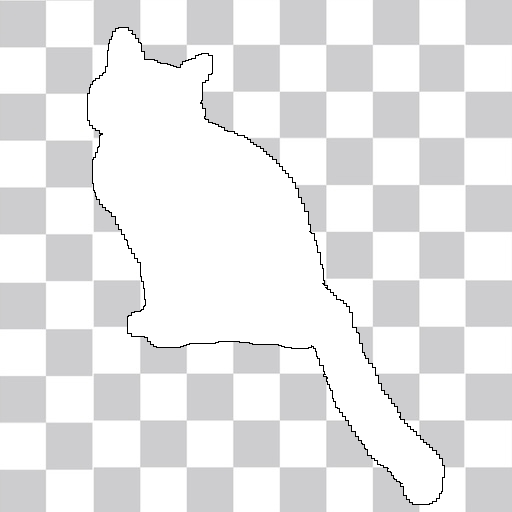} 
    \includegraphics[width=0.22\linewidth]{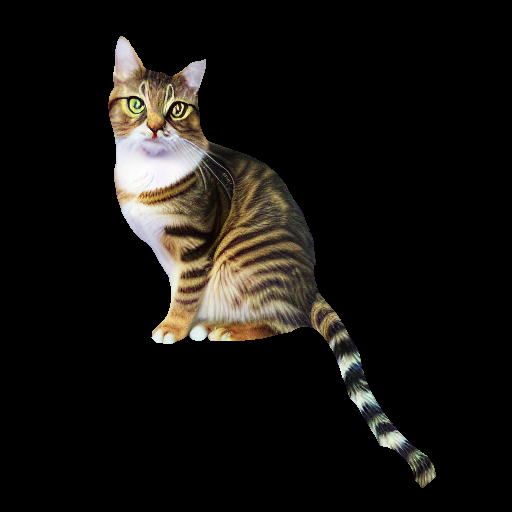} 
    \includegraphics[width=0.22\linewidth]{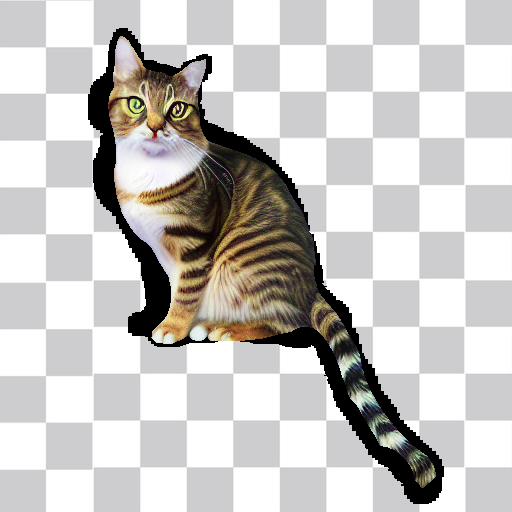} 
    \includegraphics[width=0.22\linewidth]{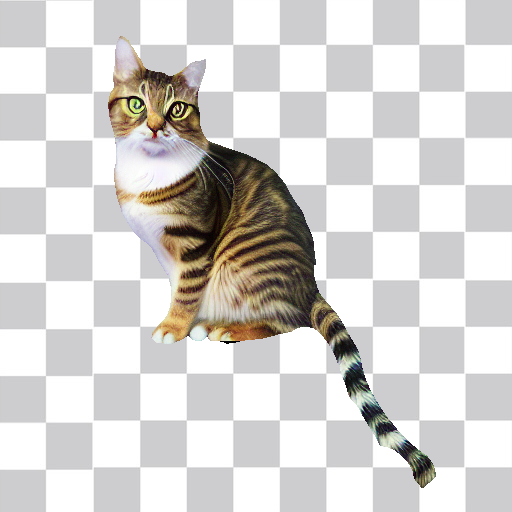} 
    \caption{Example to illustrate importance of SAM. Left to right: original mask, output of Single Object Generation, output cropped using the original mask, output cropped using the mask adjusted by SAM. }
    \label{fig:sog-example}
\end{figure}
Objects from the SOG process may not precisely follow the input mask $M_i$, especially with hand-drawn masks. 
Extracting objects using the original mask may introduce background segments in the CC stage (see Fig. \ref{fig:sog-example}). 
To address this, we perform object segmentation on the output images, 
completely separating the generated object from the image. 
Using a pre-trained Segment-Anything-Model \cite{sam} with the bounding box of the original mask as input, 
we find the intersection of the output and original masks.
\begin{equation}
    \hat M_i = \text{SAM}(x_0 | bbox(M_i)) * M_i
\end{equation}
where \( bbox(M_i)\), is the minimal bounding box spanning the mask $M_i$ (in $[x,y,w,h]$ format).

\subsubsection{Inpainting}
\label{sec:method:Inpaint}

\begin{figure}
    \centering
    \includegraphics[width=\linewidth]{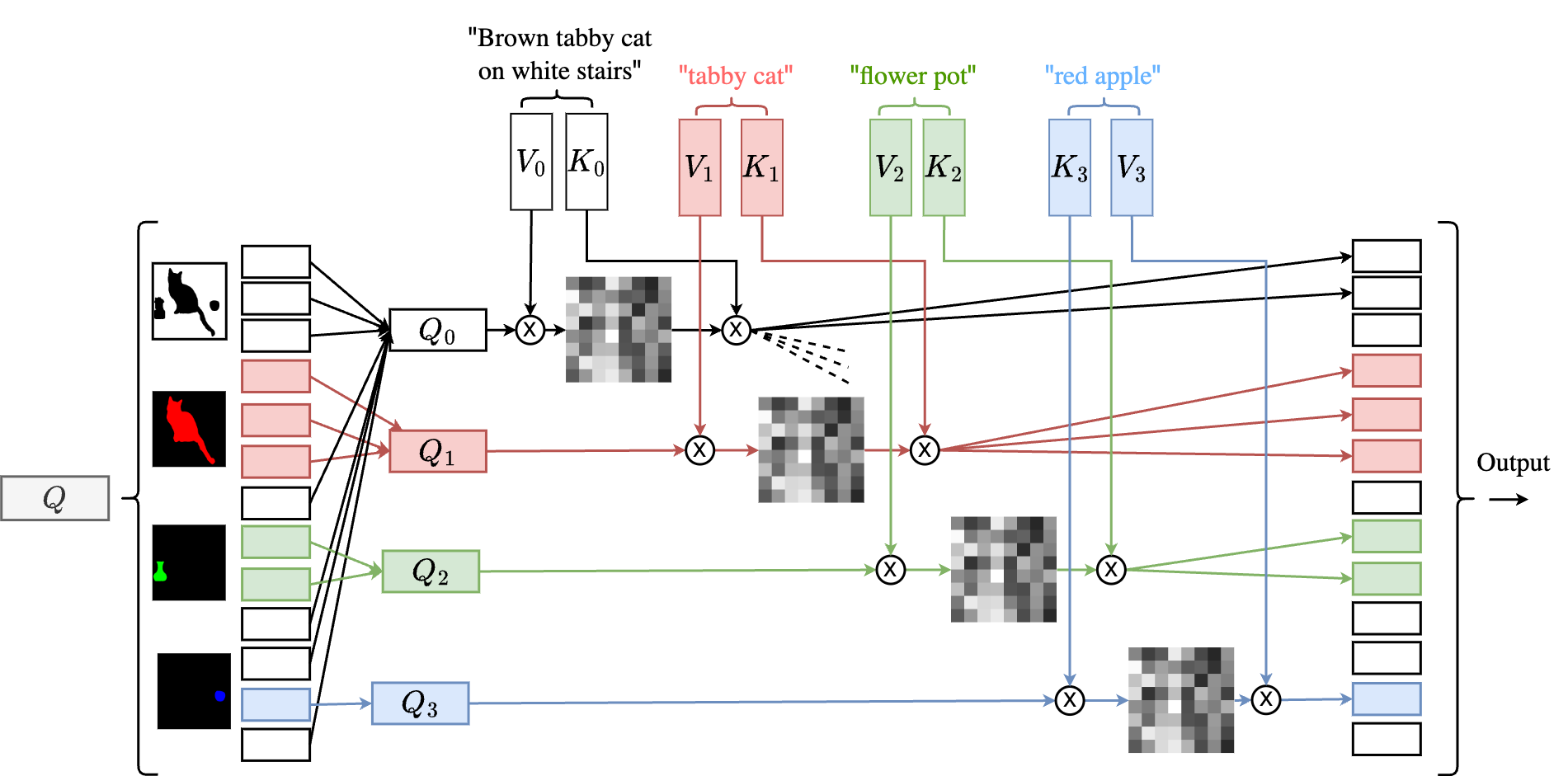} 
    \caption{Region-Grouped Cross-Attention (ReGCA) architecture.}
    \label{fig:compcomp}
\end{figure}

We combine all the components using a pre-trained Stable Inpainting model \cite{stable_diff}.
Rather than creating a new object in the masked region, we utilize the model to synthesize a background around the existing ones.
We construct the input image for the inpainting model by combining $x^i_0$ generated during SOG
using predicted masks $\hat M_i$: $x_\text{known} = \sum_i{x^i_0 \hat M_i}$.
Similarly we define the combined mask: $M = 1 - \sum_i{\hat M_i}$.

Similar to \cref{sec:method:sog}, we choose a starting step $T'' < T$, 
discouraging the model from generating new objects, and prompting it to focus on generating a background.
Additionally, to better maintain the structural coherence of pre-existing objects, 
we initialize the initial latent noise $x_{T''}$ within the known region
as the noised latent $x^\text{known}_{T''}$ of the input image $x^\text{known}$.
\begin{equation}
    x^\text{known}_{T''} = \sqrt{\alpha_{t-1}} x^\text{known} + \sqrt{1-\alpha_{t-1}} \epsilon_1
\end{equation}
\begin{equation}
    x_{T''} = M x^\text{known}_{T''} + (1 - M) \epsilon_2
\end{equation}
where $\epsilon_1$ and $\epsilon_2$ are random noise vectors sampled from $N(0,1)$

We perform DDIM iterations from timestep $T''$ down to a minimum timestep $t_\text{min}$ using the mask $M$.
For the last $t_{min}$ timesteps, we increase the mask to cover the entire image, 
allowing the model to fine-tune the known region together with the newly generated regions, 
to obtain a more homogeneous final image.

\subsubsection{Region-Grouped Cross-Attention (ReGCA)}
\label{sec:method:ReGCA}

Similar to PACA, the ReGCA (Fig. \ref{fig:compcomp}) layer is crucial for obtaining a coherent image after inpainting.
We modify cross-attention layers for two purposes: 
first, we use negative prompts to prohibit the model from generating existing objects outside the known region. 
Second, we ensure the model receives sufficient information about existing objects through cross-attention values, even when the corresponding object prompts are missing from the global caption.
To achieve this, we divide the pixels of the latent vector into groups, 
based on which object, or background they belong to (see \cref{fig:compcomp}).
For the object with index $o$, mask $\hat M_o$ and prompt $T_o$
we select the subset of queries $q_i^o = \{ q_i | i \in \hat M_o \}$.
For each group we compute its own set of key-value pairs $k_j^o$, $v_j^o$, 
as well as their unconditional counterparts $\tilde k_j^o$, $\tilde v_j^o$.
$k_j^o$, $v_j^o$ are computed from the tokens of the object prompt $T_o$,
while  $\hat K^k$ and $\hat V^k$ - using an empty string.

We add an additional group for the pixels belonging to the background $q_i^\text{bg} = \{ q_i | i \notin \hat M_o \forall o \}$.
For this group, we use the global prompt $\tau_\text{global}$ for computing $k_j^\text{bg}$ and $v_j^\text{bg}$, 
while for $\tilde k_j^\text{bg}$, $\tilde v_j^\text{bg}$ we construct a new prompt 
from the comma separated concatenation of all object prompts $T^\text{uc}_\text{bg} = \cup T_o$.
Using a non-empty prompt with the unconditional model serves as a negative prompt, preventing the generation of existing objects in the background area.

After individually computing the cross attention outputs for each group,
they are combined into a single output by putting pixels from each output
into their corresponding positions in the original input.
\section{Experiments}

\begin{figure*}[t]
    \centering
    \includegraphics[width=0.8\linewidth]{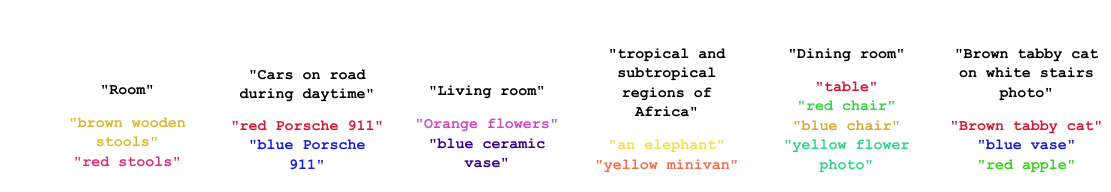}
    
    \includegraphics[width=0.8\linewidth]{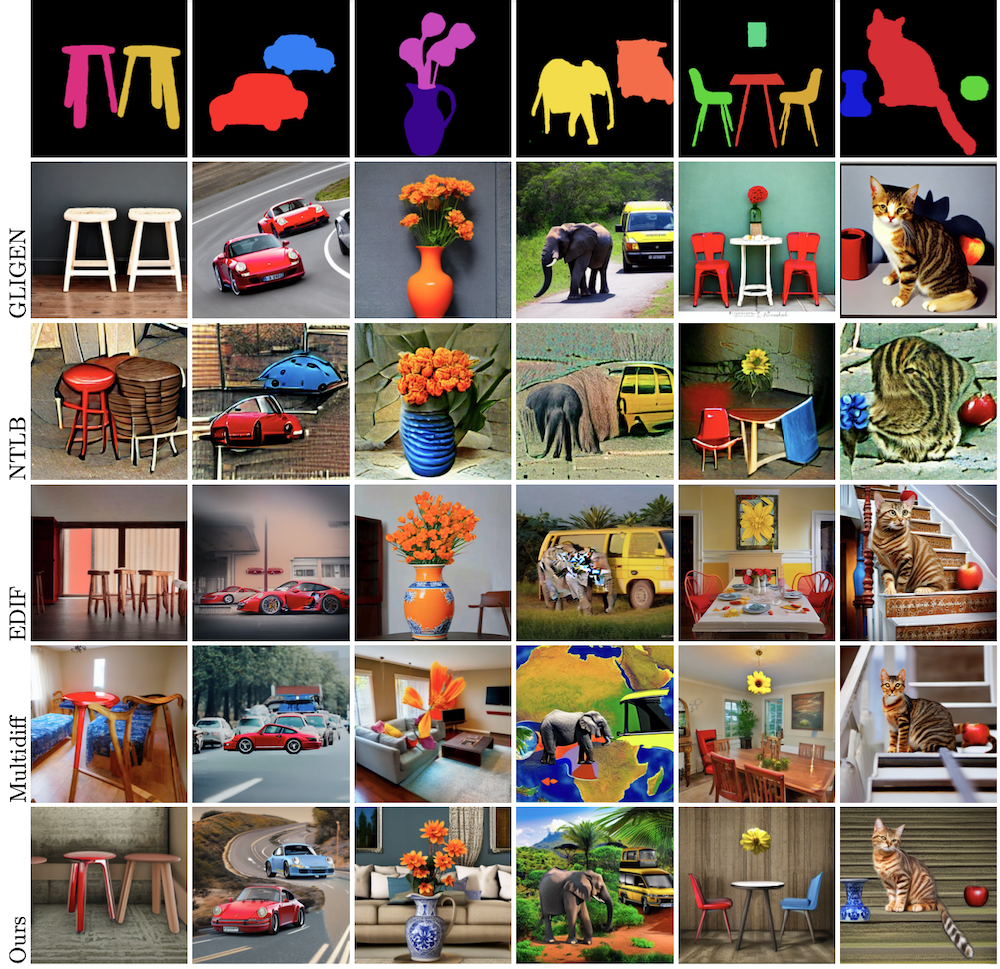}
    \caption{Qualitative comparison between the Zero-Painter pipeline and state-of-the-art models.}
    \label{fig:qualitative-comparizon}
\end{figure*}

\subsection{Implementation details}

Our implementation is based on the "Stability-AI" GitHub repository \cite{stability_ai_repo}.
We use pre-trained weights of Stable Diffusion 1.4 \cite{sd14} and Stable-Inpainting 1.5 \cite{sd15} from the huggingface repository.
For starting timesteps we use $T',T'' = 800$ and $t_{min} = 100$.
We use 40 DDIM steps (we skip 10 steps due to $T' = 800$).

\subsection{Quantitative Results}

To assess our model and compare it with other state-of-the-art models, 
we created a validation set consisting of 3000 MSCOCO \cite{cocodataset} segmentation layouts.
We filtered out masks that have an area $<5\%$ of the image, and resize all layouts to $512$x$512$.

We compared our model with existing layout2image approaches, that have publicly available repositories: 
eDiff-I using Stable Diffusion 1.4 \cite{edif_code,eDiff}, Multidiffusion \cite{multidif_code,multidiffusion},
as well as bounding-box based approaches: Gligen  \cite{gligen_code,gligen} 
and NTLB \cite{no_token_code,no_token_left_behind}. 
For the latter, we used the bounding boxes of the corresponding masks as input.
We use Local CLIP-Score to compute text-alignment: we crop the image using bounding boxes of each mask and compute the CLIP Score with the corresponding object prompt.
\begin{multline}
    \mathcal{S}_\text{CLIP}(I,C) = \\
    \frac{1}{n} \sum_{i=1}^{n} 
    \max\left(
        100 \cdot \cos\left( 
            E_\text{img} \left( I^\text{crop}_i \right), 
            E_\text{txt} (\tau_i)
        \right), 
        0 \right)
\end{multline}
where $\tau_i$ is the object prompt, $I^\text{crop}_i$ is the cropped region of image $I$ using the bounding box of $M_i$ and $E_\text{img}$, $E_\text{txt}$ correspond to extracting CLIP embedding.
We measure shape alignment using the average of Local Intersection over Union (IoU). Utilizing a pre-trained SAM model  \cite{sam} on the cropped region, we determine the shape of the final generated object and compute its IoU with the original mask.
\begin{equation}
    \mathcal{S}_\text{IoU} = 
    \frac{1}{n}\sum_{i=1}^{n} \text{IoU}(\text{SAM}(I|\text{box}(M_i)),M_i)
\end{equation}
where \text{IoU} = $\frac{\text{Area of Overlap}}{\text{Area of Union}}$ , $I$ is generated image.

As evident from \cref{tab:quantitative} Zero-Painter outperforms other state-of-the-art approaches.

\subsection{Qualitative Results}
To present a more detailed and visual comparison of our model, 
we hand-crafted a smaller test-set
using images from Unsplash as the base of our layouts. For a more extensive comparison with a larger image set, see the Appendix. Meanwhile, Fig. \ref{fig:qualitative-comparizon} displays a subset of examples.

As mentioned above, the most common issue for competitor models is the leakage of properties between different objects, 
e.g. the colors of the chair in column 1, colors of the cars in column 2.
In addition, some objects can be completely neglected, like the vase in column 6 and 3 for Multidiff.
We also note, that this particular implementation of EDIFF-I might generate visual artifacts like in column 2 and 4.
Similarly, Multidiff's outputs can sometimes look "mutated", like the chairs in column 1.

In comparison, our model is less likely to struggle from all aforementioned issues. Since the objects are generated individually using PACA module and later refined with ReGCA, 
the properties of each object are much better preserved in the final image. 

\section{Ablation Study}

\begin{figure}[t]
    \centering
    \includegraphics[width=1\linewidth]{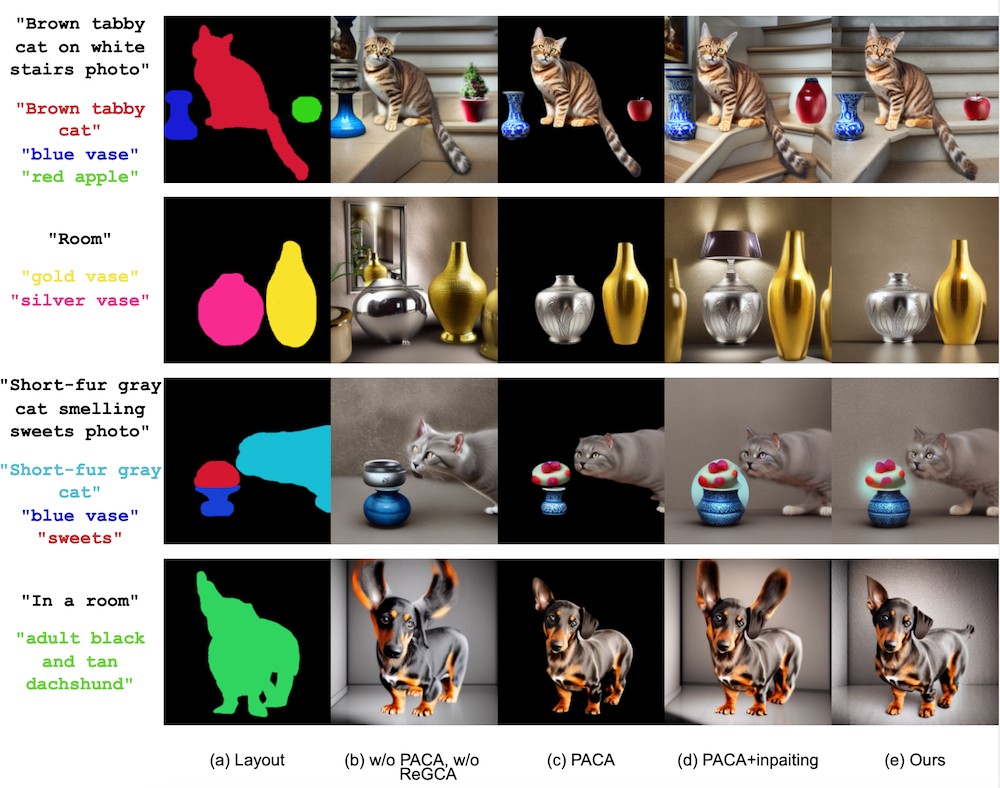}

    \caption{Ablation study.}
    \label{fig:ablation_visual}
\end{figure}

In this section, we assess the significance of two presented modules PACA and ReGCA. 
Results without PACA and ReGCA modules are presented in Fig (column 2). 
PACA ensures that the generated object tries to fill as much of the mask as possible Fig (column 3). 
Moreover by increasing the similarity with all tokens in the object prompt, 
PACA ensures that the object properties are kept as closely as possible.
For instance, consider the red apple in the 1st row, 3rd column, which is generated using PACA. 
Notably, in the 1st row, 2nd column, where "red apple" is intended, the model generated something similar  to a "red vase".
The impact of PACA becomes more evident in the 3rd row, where, in its absence, 
the model generates a gray blob instead of the intended "candies." 
This deviation highlights PACA's crucial role in maintaining coherence between generated content and textual description.

The CC stage faces challenges even when objects are accurately generated in the PACA stage. 
When using a basic inpainting pipeline, two issues arise. 
First, some objects extend beyond their intended boundaries, 
disrupting visual coherence (column 4,  see the dog's ears in row 4 and the vase in the 1st row). 
Second, inpainting with $\tau_\text{global}$ fails when it lacks information about all objects, 
as seen in the first row where the absence of the "apple" prompt results in a failed inpainting. 
ReGCA in column 5 prevents shape continuation, addressing limitations due to missing object prompts in $\tau_\text{global}$.
\section{Limitations}
\begin{table}[]
    \setlength{\tabcolsep}{4pt}
    \small
    \centering
    \begin{tabular}{c|c|c|c|c|c}
        \textbf{Model} & 
        \makecell{\textbf{EDIFF-I}} & 
        \makecell{\textbf{GLIGEN}} & 
        \makecell{\textbf{NTLB}} &
        \makecell{\textbf{MDF}} & 
        \makecell{\textbf{Ours}} \\
        \hline
        \hline
        \textit{\makecell{CLIP (local)}} & 25.3 & 25.52 & 25.71 & \underline{26.10} & \textbf{26.68} \\
        \textit{\makecell{IoU (local)}} & \underline{0.62} & N/A & 0.50 & 0.58 & \textbf{0.75} \\
    \end{tabular}    
    \caption{Quantitative comparison.}
    \label{tab:quantitative}
\end{table}
\begin{figure}[t]
    \centering
    \includegraphics[width=1\linewidth]{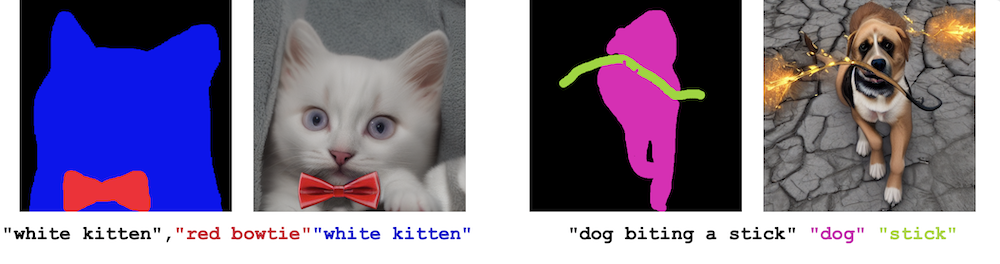}

    \caption{Zero-Painter's limitation in handling overlapping masks.}
    \label{fig:limitations}
\end{figure}

While Zero-Painter excels in generating detailed and controlled images, there are still limitations. 
One such limitation arises when dealing with overlapping masks. 
In instances where masks intersect or overlap, the resulting images may exhibit unnatural or less visually coherent outcomes (see Fig. \ref{fig:limitations}).
This challenge occurs during the CC stage: 
since we are using ReGCA for inpainting only the background region, objects inside masks remain unchanged.
Although Zero-Painter excels in many scenarios, improving this limitation is an area for future enhancement.
\section{Conclusion}

To this end, we propose Zero-Painter that is a training-free framework for layout-conditional text-to-image synthesis. 
The method utilizes object masks, individual descriptions, and a global text prompt, 
employing a two-stage process with novel Prompt-Adjusted Cross-Attention (PACA) and Region-Grouped Cross-Attention (ReGCA) blocks. 
These advancements ensure precise alignment of generated objects with textual prompts and mask shapes. 
Extensive experiments demonstrate that Zero-Painter's ability to preserve textual details and keeping shapes are superior to all existing methods.
The paper introduces innovative contributions, including the novel framework, PACA and ReGCA blocks, and comprehensive experimental validation.
\clearpage

{
    \small
    \bibliographystyle{ieeenat_fullname}
    \bibliography{main}
}

\clearpage
\setcounter{page}{1}
\maketitlesupplementary
\appendix

\section{Additional Discussion About the Single Object Generation Stage}

\begin{figure}
    \centering
    \includegraphics[width=0.8\linewidth]{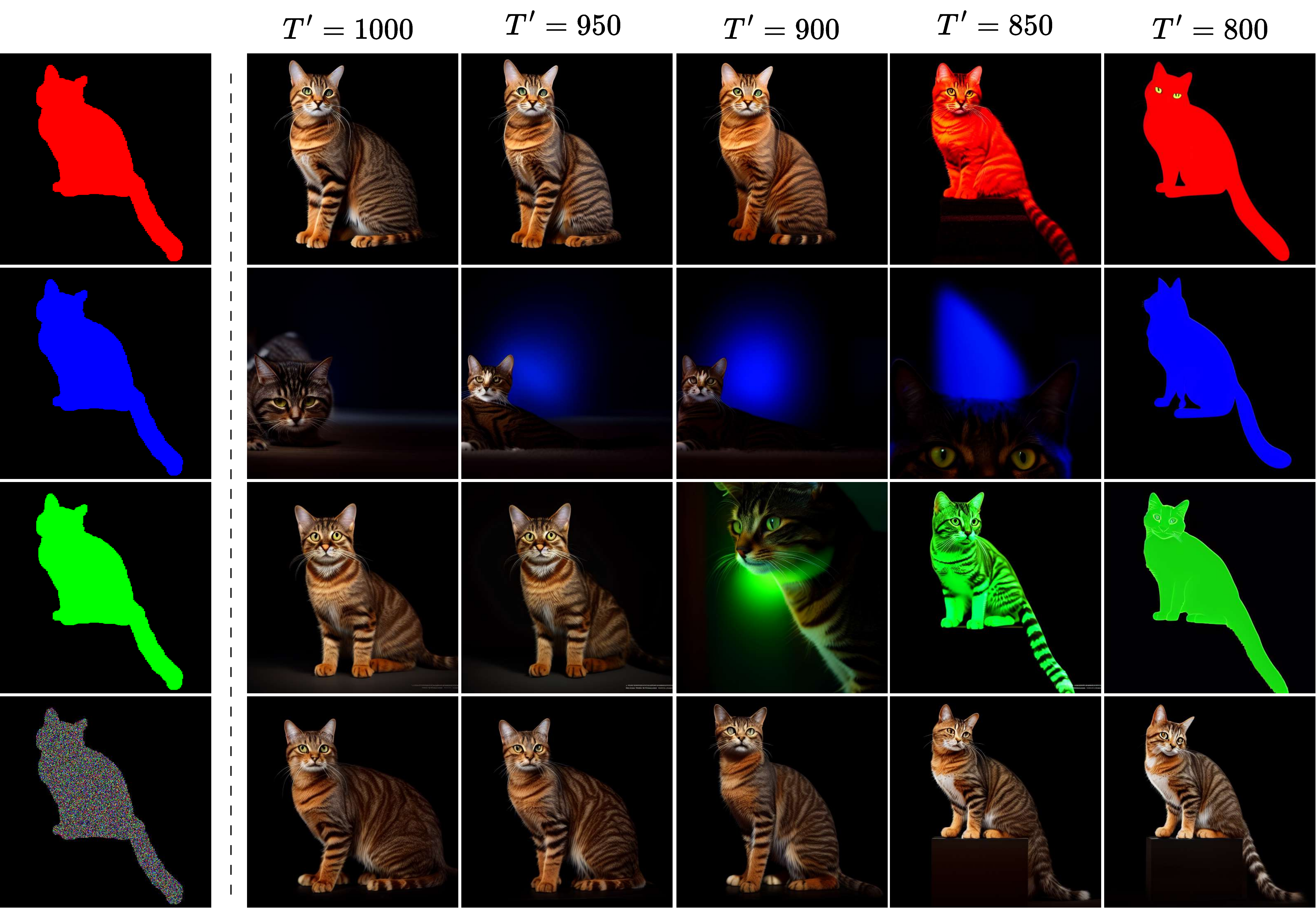}
    \caption{The effect using a starting latent with a specific color in \cref{eq:method:starting_latent} compared to initializing it with a random $\epsilon$. For each color the images show the corresponding final output when the generation starts at timestep $T'$}
    \label{fig:sup:starting-color}
\end{figure}

\begin{figure}
    \centering
    \includegraphics[width=0.5\linewidth]{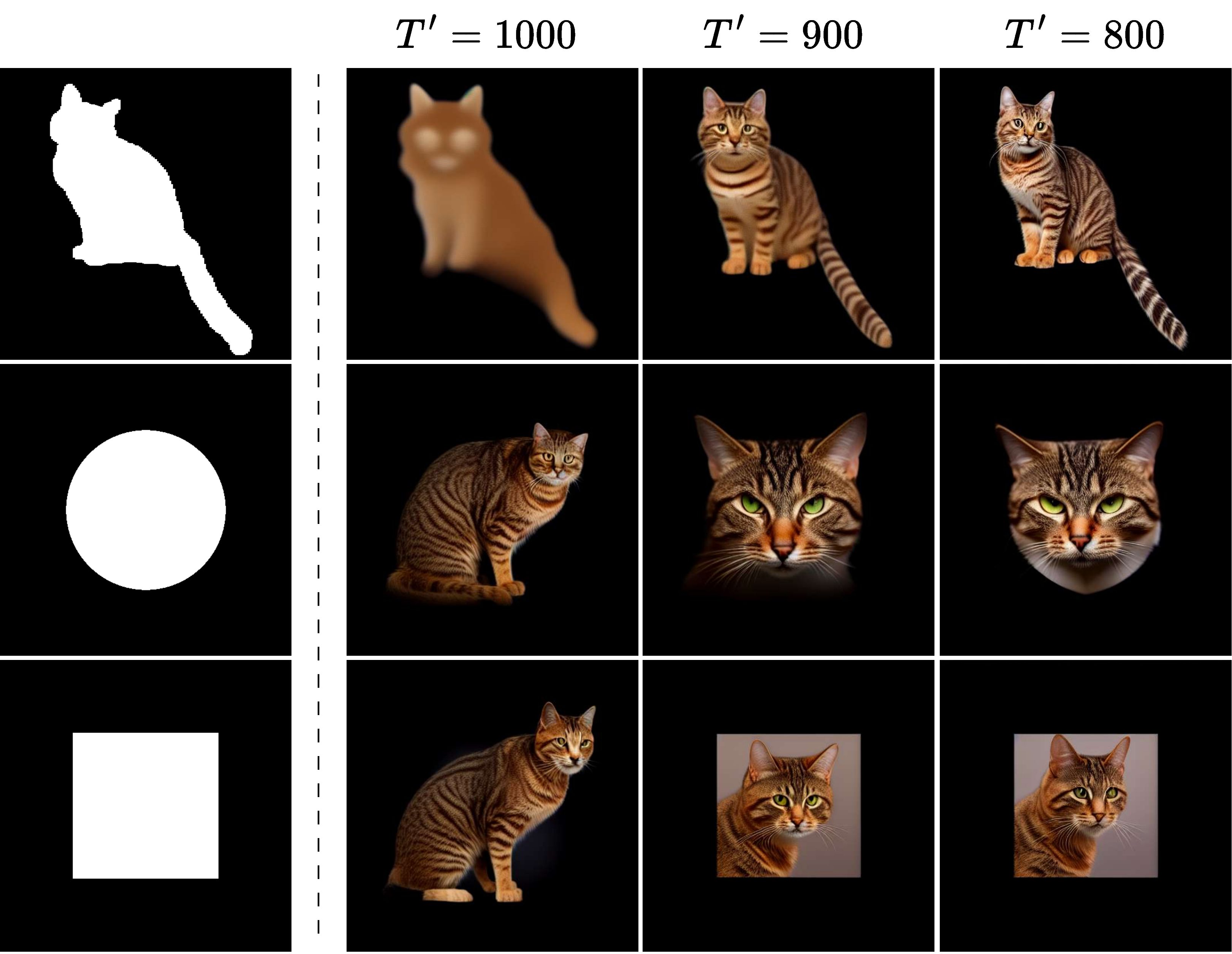}
    \caption{The effect of specifying the starting shape in \cref{eq:method:starting_latent}. The images show the corresponding final output when the starting latent is initialized with a random $\epsilon$ as per \cref{eq:method:starting_latent} and the generation starts at timestep $T'$. Apart from the starting latent noise and the starting timestep the DDIM process and the Unet have not been modified (no PACA, ReGCA, etc).}
    \label{fig:sup:starting-shape}
\end{figure}

\begin{figure}
    \centering
    \includegraphics[width=1.0\linewidth]{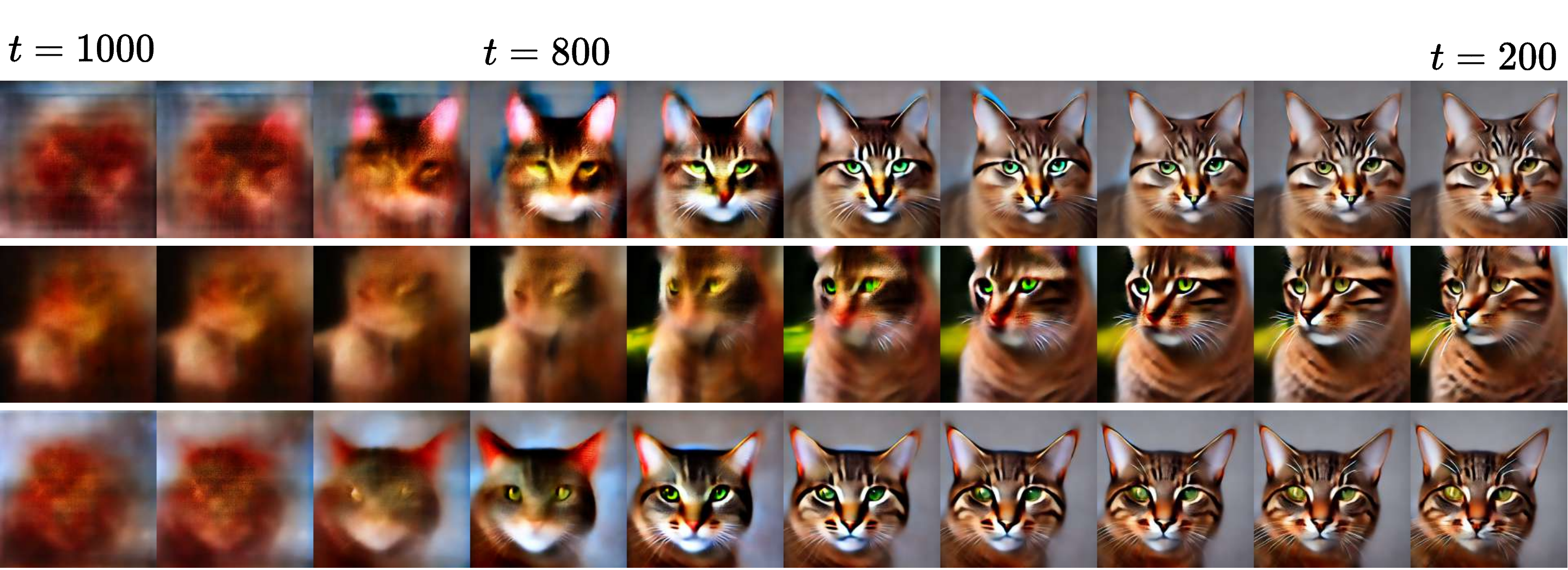}
    \caption{Illustration of the generation process of a "cat" image with DDIM. Early timesteps focus on generating rough silhouettes, while later steps generate details. Images show the predicted $x_0$ at the corresponding timestep $t$}
    \label{fig:sup:timestep-start}
\end{figure}

As mentioned earlier, \cite{eDiff} demonstrated that early generation stages focus on rough silhouettes and shapes, while later stages concentrate on details. 
This is illustrated in \cref{fig:sup:timestep-start}. 
In \cref{fig:sup:timestep-start}, during the DDIM process, we decode the predicted $x_0$ at intermediate timesteps $t$ and present the results. 
Our hypothesis is that skipping early timesteps, while providing some information about the shape, will cause the model to stick to that shape regardless of the prompt. 
We demonstrate this claim in \cref{fig:sup:starting-shape}. 
We initialize the starting latent using the same algorithm described in the paper, 
employing a latent black color for the background and a Gaussian for the masked region, 
but we omit the intermediate blending step from the generation process.
Additionally, we eliminate the PACA layer modification,  
meaning that, apart from the starting noise initialization, there is nothing restricting the model
from generating non-zero pixels outside of the masked area.
As can be seen, while the model exceeds the masked are for high values of $T'$,
for smaller values it sticks to the input shape quite closely.

Finally, in \cref{sec:method:sog}, we mention that providing specific color information to the generation will influence the final output of the model, which is undesirable.
This effect is illustrated in \cref{fig:sup:starting-color}.
Here, we initialize the background using the latent black color, and for the masked area, 
we use the same \cref{eq:method:flat_latent} but with a mask of the specified color instead of black.
As can be seen, while for high $T'$ the influence on the output color is minimal, 
for smaller $T'$ the influence grows.
For $T' = 800$, which is particularly convenient for shape-awareness, the color influence is particularly high. 
In contrast, with the random initialization introduced in our paper, 
the output image exhibits much more organic colors, while the quality loss of the final image is minimal.

\section{User Study}

\begin{figure*}
    \centering
    \includegraphics[width=0.49\linewidth]{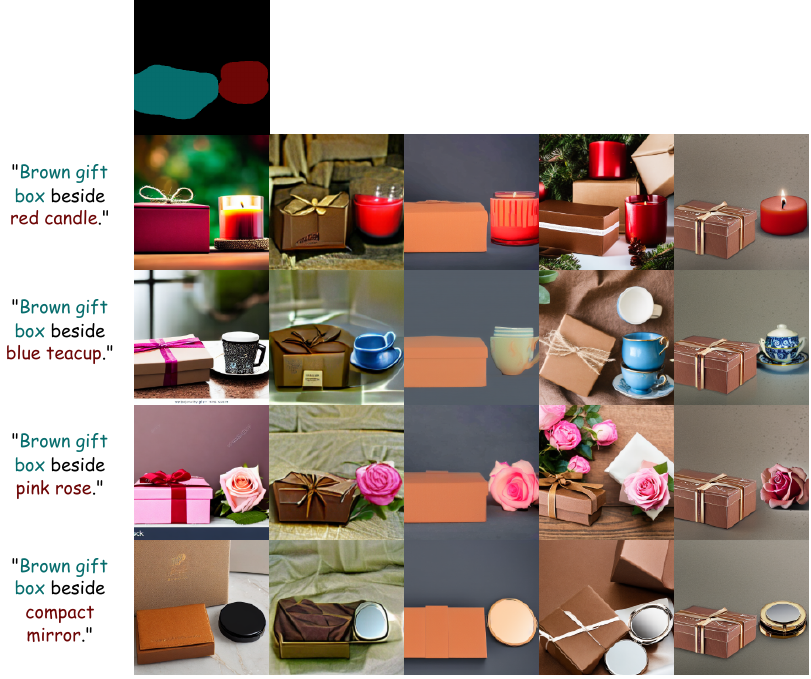}%
    \includegraphics[width=0.49\linewidth]{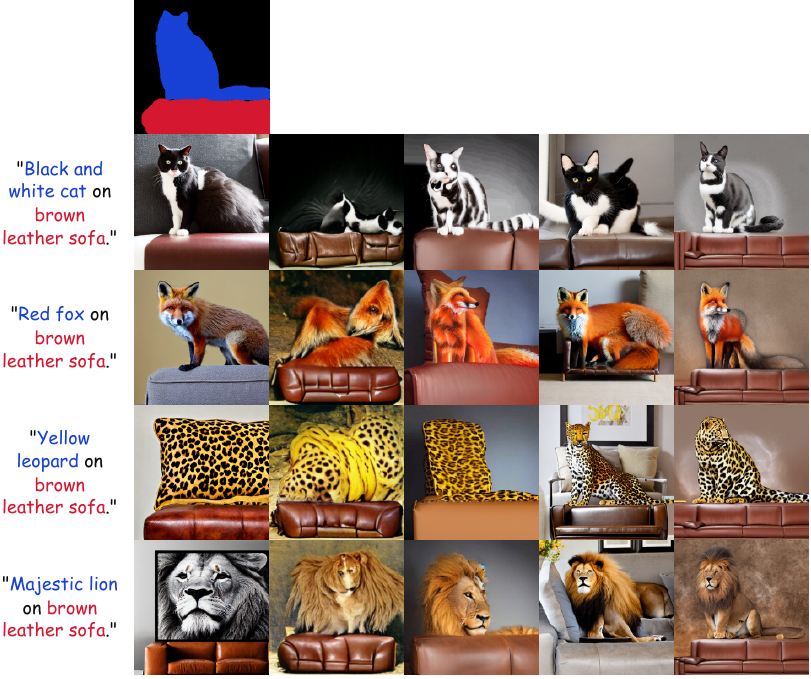}
    {
        \fontsize{8}{12}\selectfont
        \begin{tabular}{cccccccccccc}
           \hspace{0.8cm} & GLIGEN & NTLB & EDIFF-I & MULTIDIFF & OURS 
           \hspace{1.4cm} & GLIGEN & NTLB & EDIFF-I & MULTIDIFF & OURS 
        \end{tabular}
    }
    \caption{Independent Object Modification: Thanks to the Single-Object Generation stage, which assigns separate object seeds for the objects in the layout, our method enables the modification of some objects while keeping the rest the same.}
    \label{fig:sup:diversity}
\end{figure*}

\begin{figure*}
    \centering
    \includegraphics[width=0.9\linewidth]{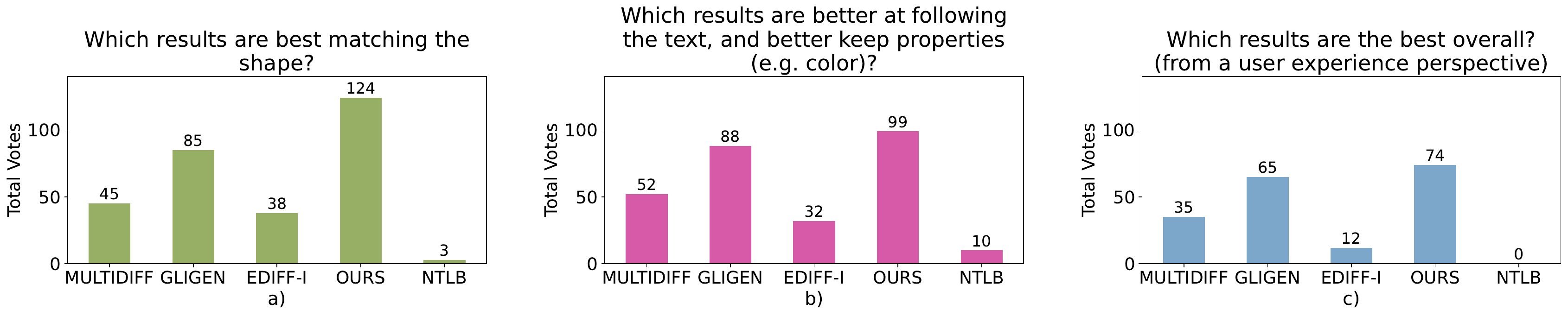}
    \caption{
        Total votes of each method based on our user study for questions a) \textit{Which results are better at following the text, and better keep properties (e.g. color)?}  b) \textit{Which results are best matching the shape?} and c) \textit{Which results are the best overall? (from a user experience perspective)}
        The user study shows a clear advantage of our method.
    }
    \label{fig:user_study}
\end{figure*}

In addition to our numerical evaluations we conducted a user study to obtain a more-comprehensive and user-oriented evaluation of our results.
The study involved 10 participants, that were tasked with evaluating our chosen competitor methods:
eDiff-I using Stable Diffusion 1.4\cite{edif_code,eDiff}, Multidiffusion\cite{multidif_code,multidiffusion},
Gligen \cite{gligen_code,gligen} and NTLB\cite{no_token_code,no_token_left_behind},
as well as our own method.
We presented the participants with 20 samples consisting of the input layout, 
the global prompt and the set of per-object local prompts, 
as well as 5 generated images - one per method, which were shuffled in a random order.
Subsequently, the participants were asked the following questions:

\begin{enumerate}[label=(\alph*)]
    \item \textit{Which results are best at matching the shape?}
    \item \textit{Which results are better at following the text, and better keep properties (e.g. color)?}
    \item \textit{Which results are the best overall? (from a user experience perspective)}
\end{enumerate}

The participants were allowed to choose none or multiple best methods since some approaches can perform equally good or equally bad.
After collecting the feedback we calculate the sum of the votes for each method across all 20 samples and 10 participants (200 points max).
The results are presented in Fig. \ref{fig:user_study} demonstrating a clear advantage of our method in both aspects: 
shape alignment and characteristic preservation of the generated objects.

\section{Quantitative Ablation Study}
In the main paper, we assessed the significance of two presented modules, PACA and ReGCA, using visual examples. Please see Table \ref{tab:ablation_study_supp} for a qualitative comparison, showcasing the importance of PACA and ReGCA.

\setlength{\belowcaptionskip}{-18pt}
\begin{table}[]
    \renewcommand{\arraystretch}{1}
    \setlength{\tabcolsep}{10pt}
    \tiny
    \centering
    \begin{tabular}{cccc}
        \toprule
        \textbf{Model} & 
        \makecell{\textbf{w/o PACA,}\\ \textbf{ w/o ReGCA}} & 
        \makecell{\textbf{PACA+} \\ \textbf{inpaiting}} & 
        \makecell{\textbf{Ours}} \\
        \midrule
        \textit{\makecell{CLIP (local)}} & 26.3 & 26.5 & \textbf{26.63} \\
        \textit{\makecell{IoU (local)}} & 0.73 & 0.74 & \textbf{0.75} \\
        \bottomrule
    \end{tabular}    
    \caption{Quantitative ablation study.}
    \label{tab:ablation_study_supp}    
\end{table}

\section{Inference speed}

\begin{figure}
    \tiny
    \centering
    \includegraphics[width=\linewidth]{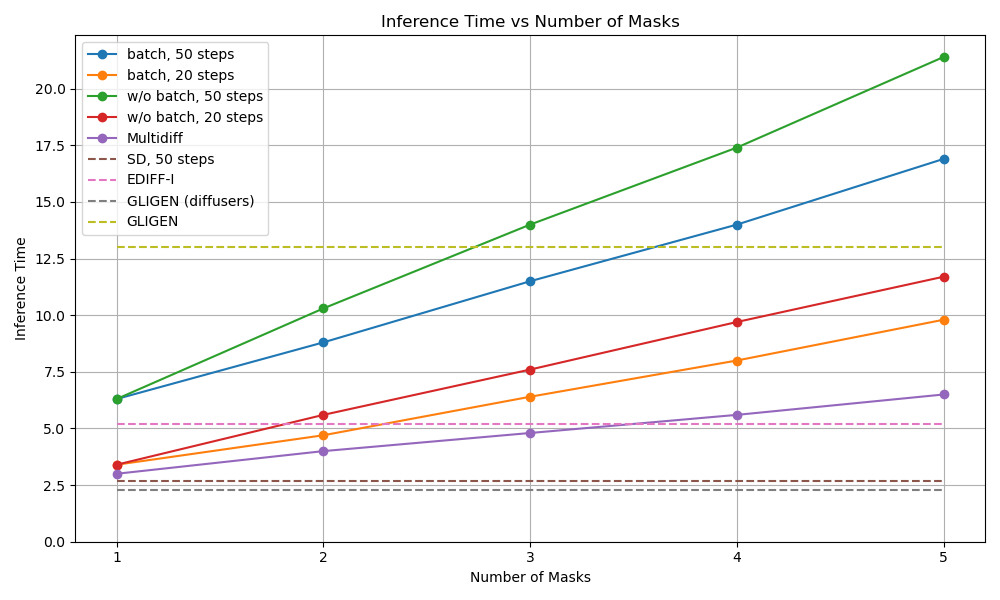}
       
    \caption{Inference time on A100 GPU for 512x512 resolution}
    \label{fig:reb:limitations_and_time}
\end{figure}

Zero-Painter's inference time is directly influenced by the number of objects generated individually during the SOG stage. This relationship is illustrated in Fig. \ref{fig:reb:limitations_and_time}, where we provide a runtime analysis and compare it with other methods. Notably, employing a batch approach for object generation results in a lower computational overhead. Moreover, as existing objects undergo refinement in the later stages of the Comprehensive Composition phase, it becomes feasible to reduce the number of diffusion steps during SOG. This reduction contributes to a decrease in the overall generation time without compromising the final output quality significantly.

\section{Additional Comparisons}

\begin{figure*}
    \centering
    \includegraphics[width=0.49\linewidth]{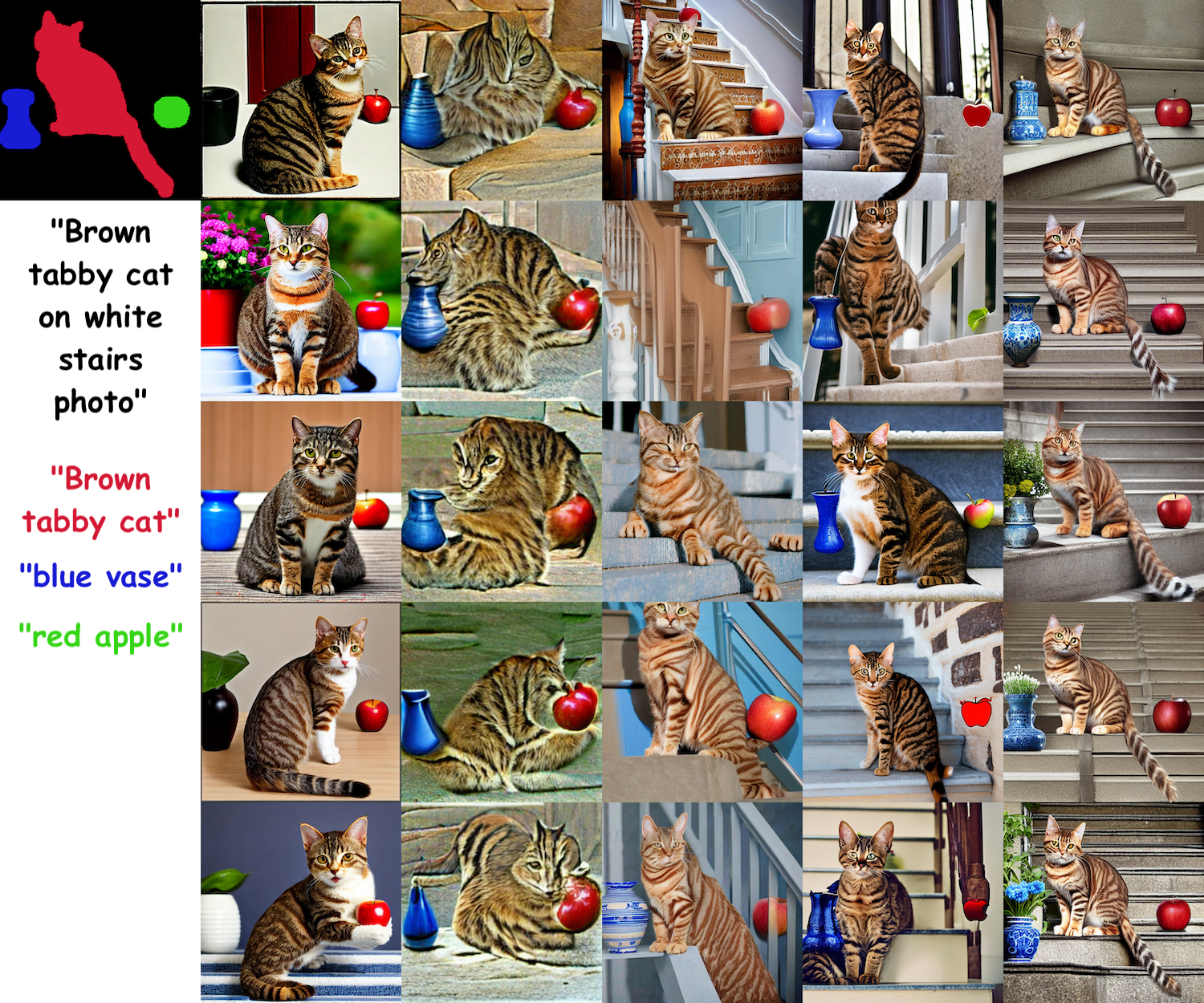}%
    \includegraphics[width=0.49\linewidth]{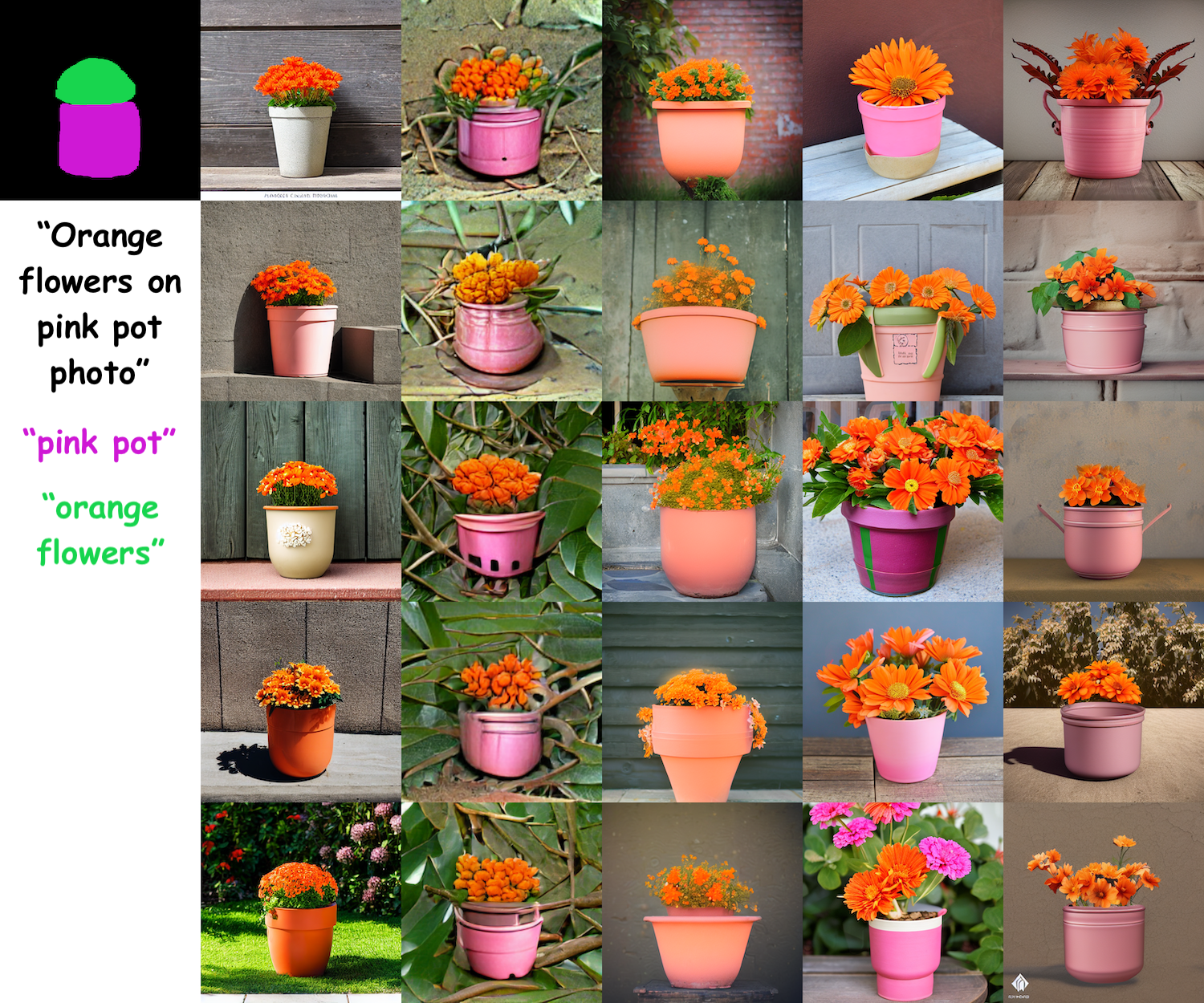}
    \includegraphics[width=0.49\linewidth]{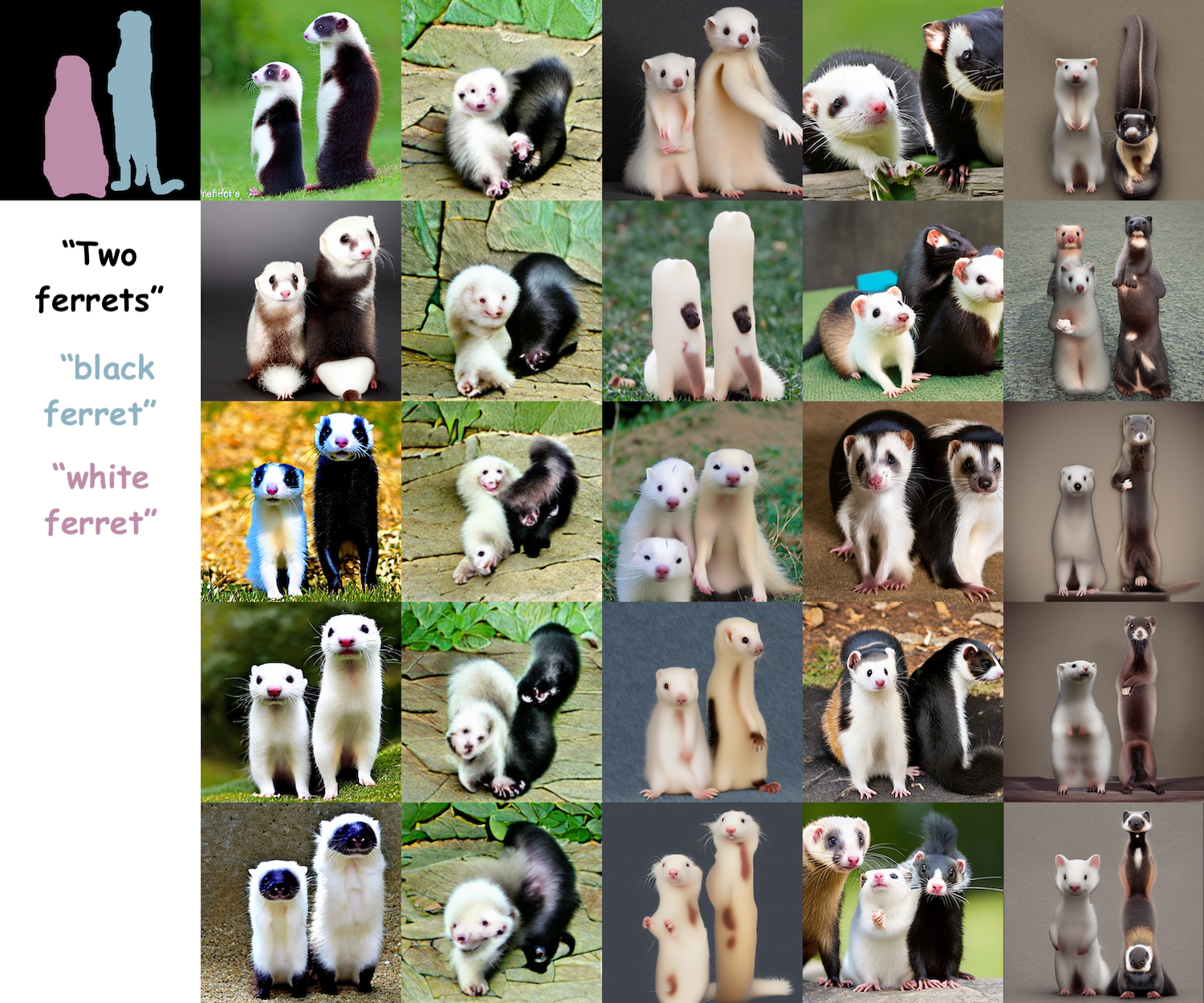}%
    \includegraphics[width=0.49\linewidth]{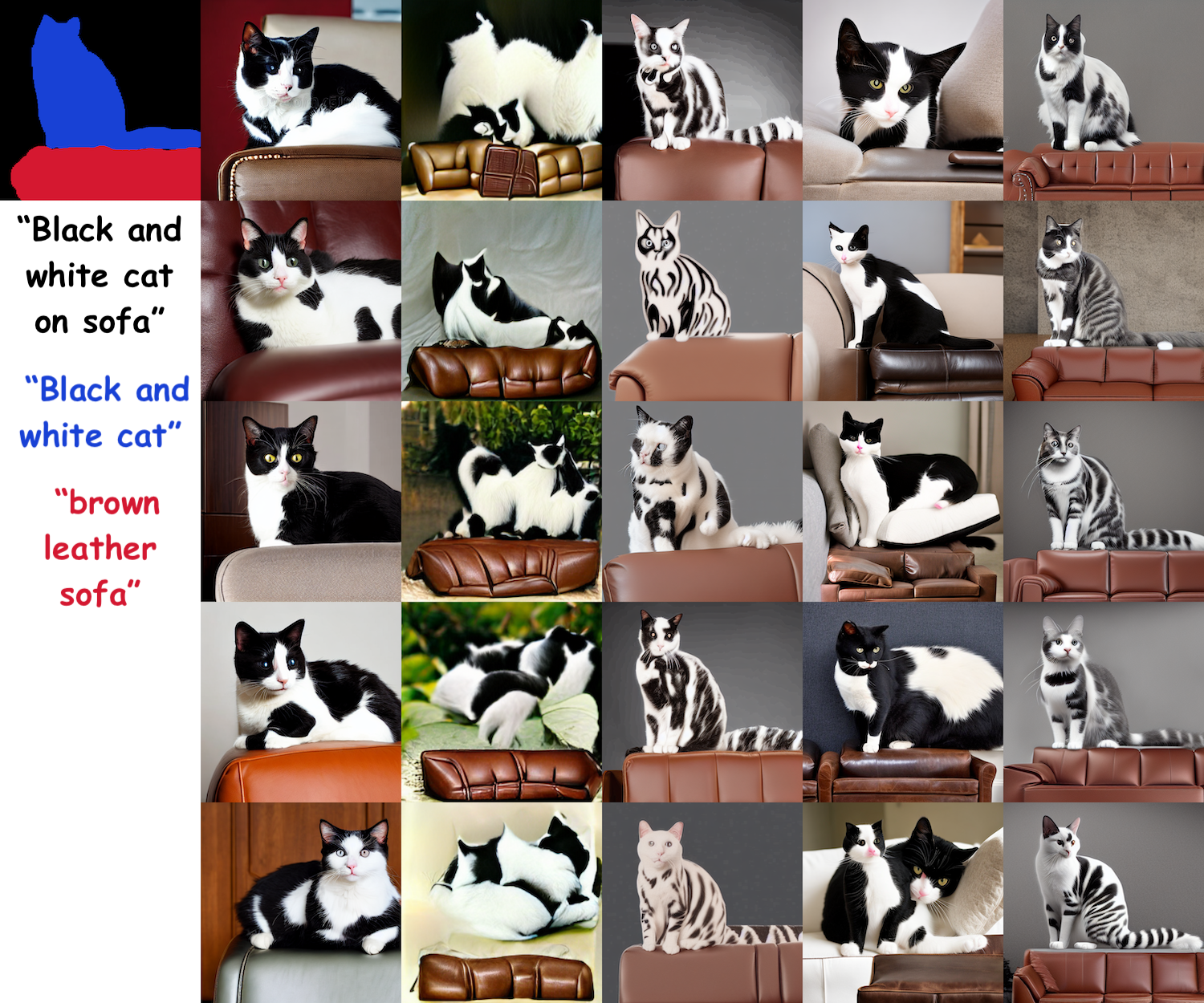}
    \includegraphics[width=0.49\linewidth]{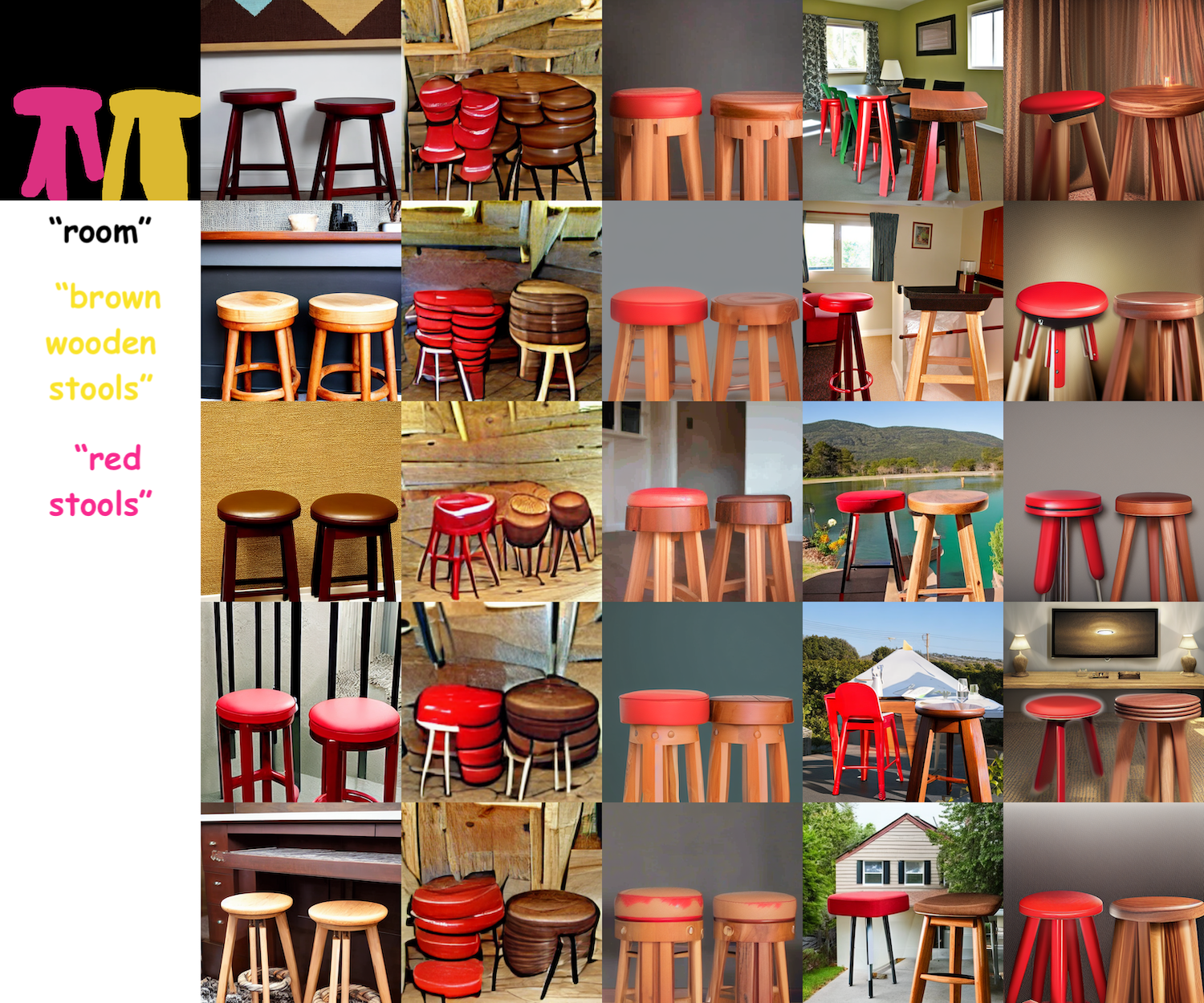}%
    \includegraphics[width=0.49\linewidth]{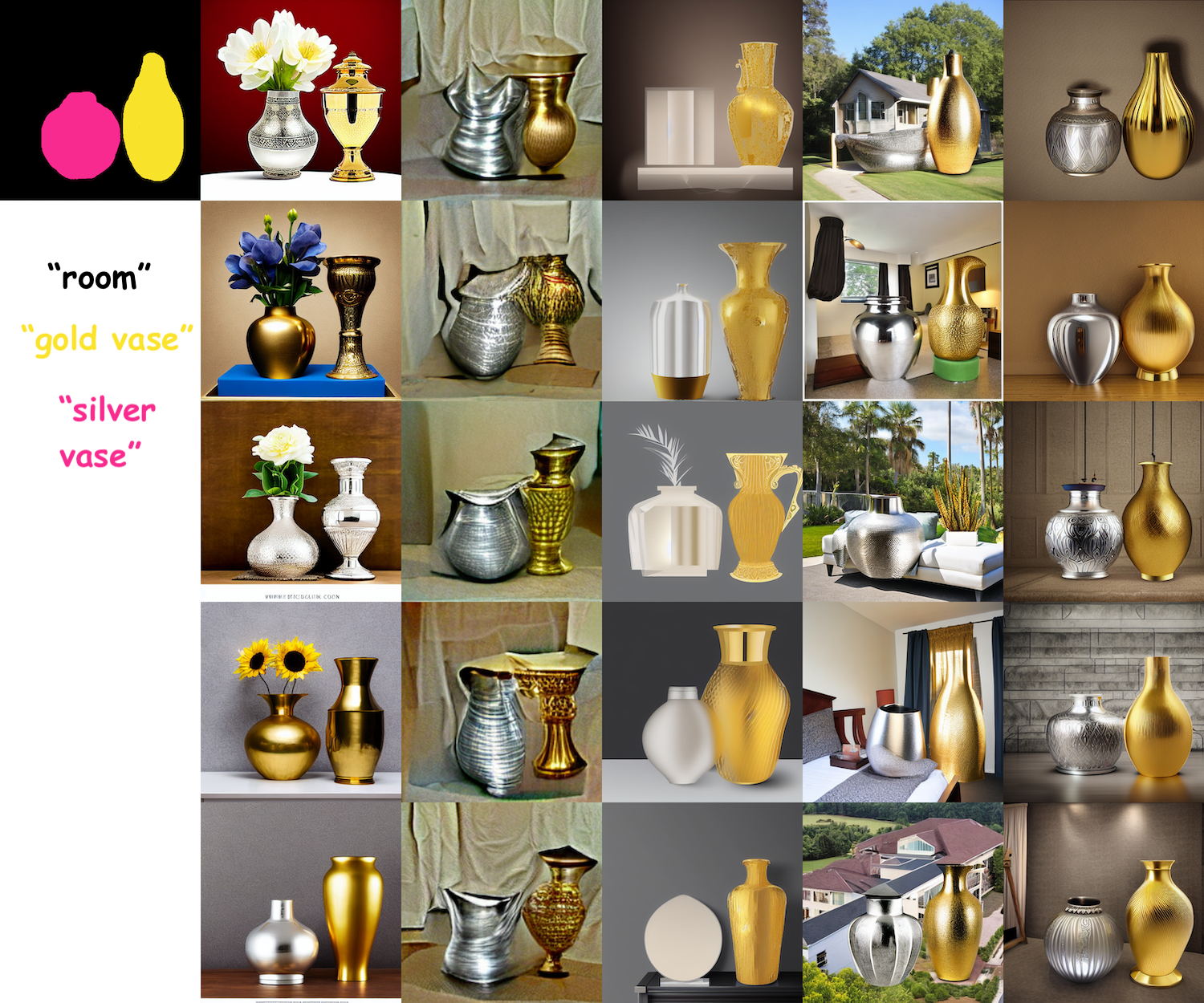}
    {
        \fontsize{8}{12}\selectfont
        \begin{tabular}{cccccccccccc}
           \hspace{0.8cm} & GLIGEN & NTLB & EDIFF-I & MULTIDIFF & OURS 
           \hspace{1.4cm} & GLIGEN & NTLB & EDIFF-I & MULTIDIFF & OURS 
        \end{tabular}
    }
    \caption{A comprehensive comparison of our method to the competitors. For each image we randomly sampled 5 seeds and used all the results. }
    \label{fig:sup:comp-random}
\end{figure*}

Since most of the compared methods, including ours, are stochastic, 
we conducted additional experiments to demonstrate that the improvements we presented are not accidental but rather consistent.
To do this, we sample five random seeds for each image in our visual test set in advance.
We generate both our and competitors' images using the same set of seeds, and show the results in \cref{fig:sup:comp-random}.
As observed, the results of our method are consistently better aligned with both the input prompt and the shape when compared to all the competitors'.

\section{Independent Object Modification}

During the Single-Object Generation stage, different objects in the image are generated separately and can use their own independent generation seed. 
Additionally, the Comprehensive Composition stage does not modify the pre-existing objects too much. This adds increased flexibility to our model. Unlike the competitors, where changing a single word in the prompt can result in dramatically different output images, our model allows changing one or more of the input objects while keeping all the others the same \cref{fig:sup:diversity}.

\begin{figure*}
    \centering
    \includegraphics[width=0.8\linewidth]{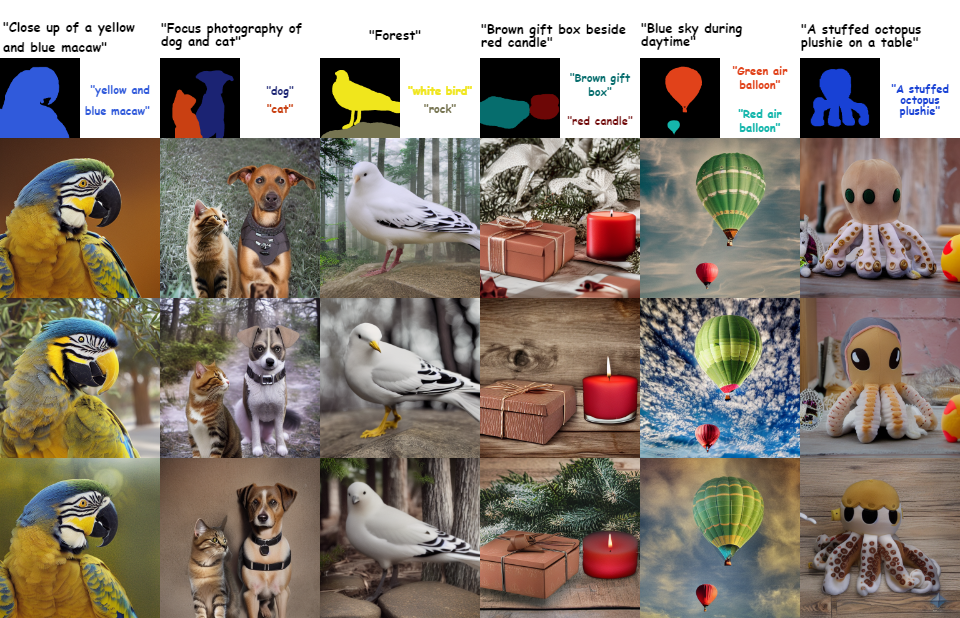}
    \includegraphics[width=0.8\linewidth]{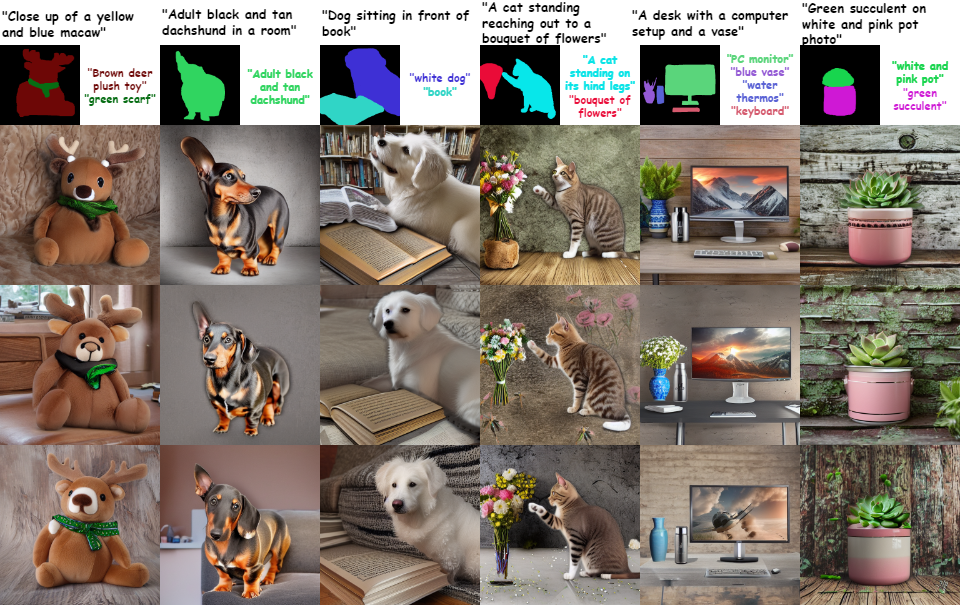}
    \caption{Additional examples of images generated by our model.}
    \label{fig:sup:additional}
\end{figure*}

\end{document}